\documentclass[lettersize,journal]{IEEEtran}

\usepackage[accsupp]{axessibility}
\usepackage[misc]{ifsym}
\usepackage[super]{nth}
\usepackage{amsmath,amsfonts}
\usepackage{array}
\usepackage[caption=false,font=normalsize,labelfont=sf,textfont=sf]{subfig}
\usepackage{textcomp}
\usepackage{stfloats}
\usepackage{url}
\usepackage{verbatim}
\usepackage{graphicx}
\usepackage{multirow}
\usepackage{colortbl}
\usepackage{bm}
\usepackage[ruled,vlined]{algorithm2e}
\usepackage{booktabs}
\usepackage{mathtools}
\usepackage{textcomp}
\usepackage{cite}
\usepackage[pagebackref,breaklinks,colorlinks,citecolor=blue]{hyperref}
\usepackage{graphicx}
\usepackage{etoolbox}
\usepackage{caption}

\usepackage[dvipsnames]{xcolor}

\newcommand{\final}{1}
\usepackage{graphicx}
\usepackage{amsmath}
\usepackage{amssymb}
\usepackage{booktabs}
\usepackage[utf8]{inputenc} 
\usepackage[T1]{fontenc}    
\usepackage{url}            
\usepackage{booktabs}       
\usepackage{amsfonts}       
\usepackage{nicefrac}       
\usepackage{microtype}      
\usepackage{xcolor}         
\usepackage{graphicx}
\usepackage{amsmath}
\usepackage{amssymb}
\usepackage{makecell}
\usepackage{engord}
\usepackage[normalem]{ulem}
\usepackage{adjustbox}
\usepackage{enumitem}

\usepackage{ifthen}

\definecolor{FanColor}{rgb}{0.8,0,0.8}
\newcommand{\fan}[1]{{\color{FanColor}[Fan: #1]}}

\definecolor{XingjiaColor}{rgb}{0.0,0.1,0.9}
\newcommand{\xingjia}[1]{{\color{XingjiaColor}[Xingjia: #1]}}

\newcommand{\warning}[1]{{\it\color{red} #1}}
\newcommand{\toremove}[1]{{\it\color{red} (To remove) #1}}
\newcommand{\note}[1]{{\it\color{blue} #1}}
\newcommand{\nothing}[1]{}

\ifthenelse{\equal{\final}{1}}
{
\renewcommand{\fan}[1]{}
\renewcommand{\xingjia}[1]{}

\renewcommand{\warning}[1]{}
\renewcommand{\toremove}[1]{}
\renewcommand{\note}[1]{}
\renewcommand{\nothing}[1]{}
}
{}

\newcommand{\loss}{\mathcal{L}}
\newcommand{\contentloss}{\loss_{c}}
\newcommand{\styleloss}{\loss_{s}}

\newcommand{\inputcontent}{I_c}

\newcommand{\layernumber}{N_l}

\newcommand{\etal}{\textit{et al.}}

\usepackage[pagebackref,breaklinks,colorlinks]{hyperref}
\usepackage[capitalize]{cleveref}

\newcommand{\insertfig}{\setcounter{figure}{0}\includegraphics[width= 0.96\linewidth]{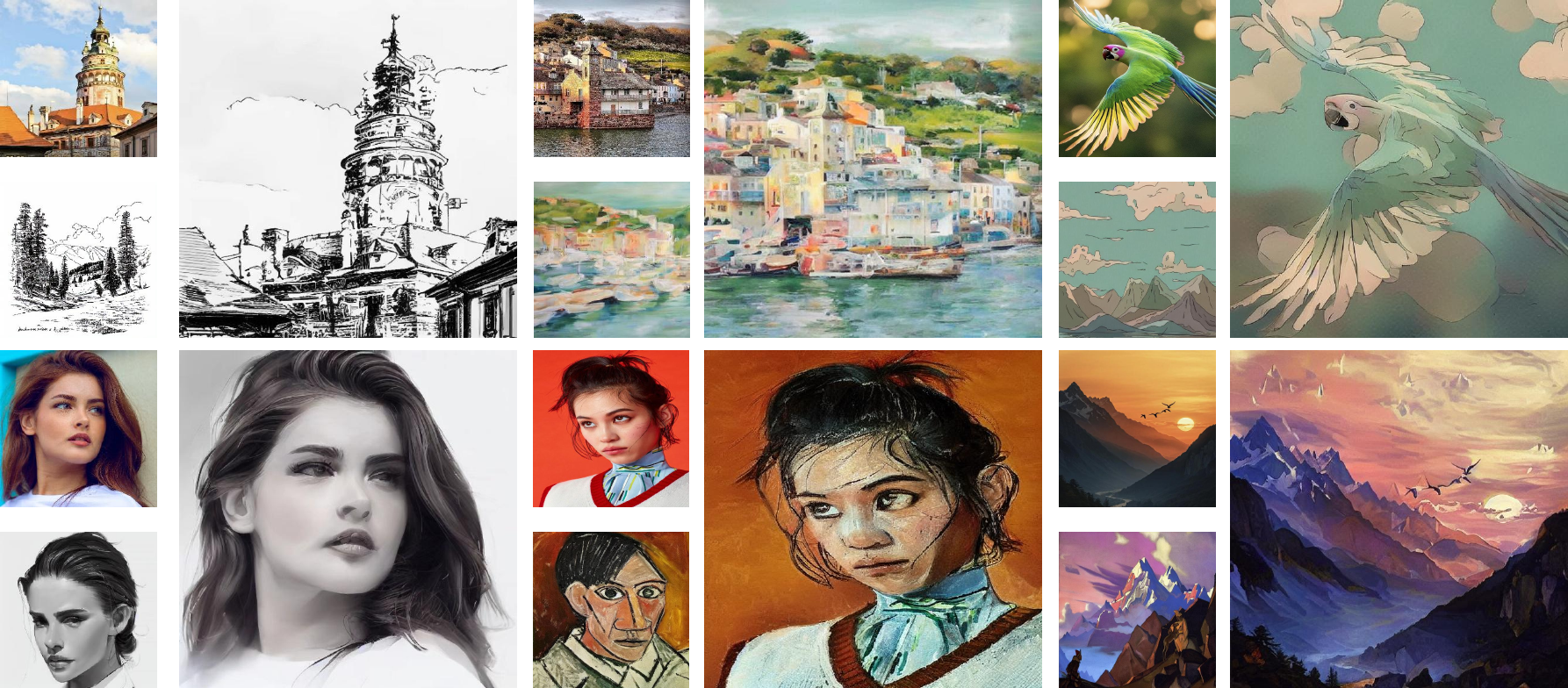}\captionof{figure}{Image style transfer results by the proposed $\mathcal{Z}$-STAR$^+$. Our method demonstrates versatility in generating stylized outputs across a diverse range of style and content reference pairs. When presented with a style image and a content image, our approach successfully produces stylized results that exhibit vibrant and distinctive style characteristics while accurately preserving the essential content details.}}

\makeatletter
\apptocmd{\@maketitle}{\centering \insertfig}{}{}
\makeatother

\begin{document}
\title{$\mathcal{Z}$-STAR$^+$: A Zero-shot Style Transfer Method via Adjusting Style Distribution}
\author{Yingying Deng,~Xiangyu He, Fan Tang, Weiming Dong
\thanks{
\IEEEcompsocthanksitem Y. Deng, X. He, and W. Dong are with Institute of Automation, Chinese Academy of Sciences, Beijing, China.  
\IEEEcompsocthanksitem F. Tang is with Institute of Computing Technology, Chinese Academy of Sciences. E-mail: tangfan@ict.ac.cn.
}

}

\markboth{Preprint}%
{Deng \MakeLowercase{\textit{et al.}}: $\mathcal{Z}$-STAR$^+$: A Zero-shot Style Transfer Method via Adjusting Style Distribution}

\maketitle

\begin{abstract}

Style transfer presents a significant challenge, primarily centered on identifying an appropriate style representation. Conventional methods employ style loss, derived from second-order statistics or contrastive learning, to constrain style representation in the stylized result. However, these pre-defined style representations often limit stylistic expression, leading to artifacts.
In contrast to existing approaches, we have discovered that latent features in vanilla diffusion models inherently contain natural style and content distributions. This allows for direct extraction of style information and seamless integration of generative priors into the content image without necessitating retraining.
Our method adopts dual denoising paths to represent content and style references in latent space, subsequently guiding the content image denoising process with style latent codes. We introduce a Cross-attention Reweighting module that utilizes local content features to query style image information best suited to the input patch, thereby aligning the style distribution of the stylized results with that of the style image.
Furthermore, we design a scaled adaptive instance normalization to mitigate inconsistencies in color distribution between style and stylized images on a global scale.
Through theoretical analysis and extensive experimentation, we demonstrate the effectiveness and superiority of our diffusion-based \uline{z}ero-shot \uline{s}tyle \uline{t}ransfer via \uline{a}djusting style dist\uline{r}ibution, termed $\mathcal{Z}$-STAR$^+$.
\end{abstract}

\section{Introduction}
\label{sec:Introduction}

\begin{figure*}
\setlength{\abovecaptionskip}{2mm}
\centering
\includegraphics[width=0.99\linewidth]{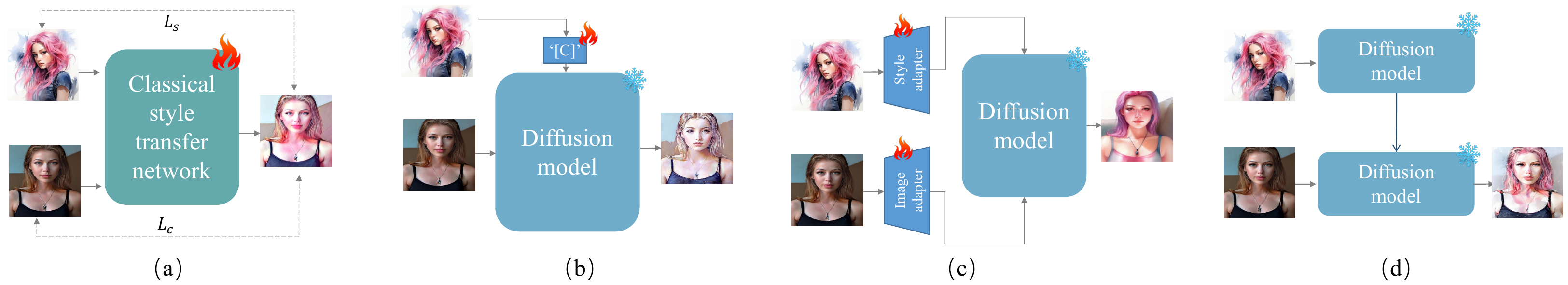}
\caption{Different pipline for style transfer task.
The schematic stylized results are generated by StyTr$^2$~\cite{Deng:2022:CVPR}, InST~\cite{zhang:2023:inversion}, InstantStyle-Plus~\cite{instancestyle} and our method.}
\label{fig:frame}
\vspace{-10pt}
\end{figure*}
\IEEEPARstart{A}{rtists} create numerous paintings formed by the unique use of strokes, color, and composition.
Migrating the style pattern of art paintings to a real image using image translation and stylization algorithm has received significant attention in the research community, with numerous machine learning techniques utilized, such as convolutional neural networks (CNN)~\cite{gatys:2016:image, Huang:2017:Arbitrary, johnson:2016:perceptual, sheng:2018:avatar, park:2019:arbitrary, deng:2020:arbitrary,DBLP:journals/tcsv/LyuJPD24, tcsvt1,tcsvt2,tcsvt3}, flow-based networks~\cite{an:2021:artflow}, visual transformers (ViT)~\cite{wu:2021:styleformer,Deng:2022:CVPR}, and diffusion models~\cite{zhang:2023:inversion,cheng:2023:ict}.

The style transfer task is inseparable from unearthing the style distribution in input images.
Gatys et al.~\cite{gatys:2016:image} propose the Gram matrix which calculates the inner product of image features to represent style.
The Gram matrix is widely used for calculating the style loss and limiting the style difference between style and result images which achieve promising results\cite{johnson:2016:perceptual, Controlling}.
Li et al.~\cite{li:2017:demystifying} treat the style transfer task as a distribution alignment of the neural activations between content and style images.
They propose a simplified style representation method by calculating the mean and variance of style features.
Certain methods, such as AdaIN~\cite{Huang:2017:Arbitrary} and InfoStyler~\cite{ DBLP:journals/tcsv/LyuJPD24} use mean and various statistics to measure global style similarity and achieve promising style transfer outcomes.
Nevertheless, the utilization of second-order statistics, such as Gram matrices or mean and variance metrics, exhibits limitations in capturing complex stylistic patterns and fails to effectively transfer localized features from the content image to the style image. In response to this challenge, CAST~\cite{zhang:2022:domain} introduces a contrastive loss function that exploits the relationships between positive and negative exemplars, thereby promoting stylistic conformity in the resultant image.
However, CAST encounters difficulties in generating stylized outputs that exhibit vivid and intricate stylistic details. 
The methods mentioned above all rely on minimizing the pre-defined style loss and optimizing network parameters, so that the network can learn style expression according to the previous definition, which leads to insufficient style rendering (e.g., hair and eyes in Fig.~\ref{fig:frame}(a)).
In the context of artistic representation, \textit{the contours and forms of a painting should be subject to the adaptable preferences of the artist's painting techniques, rather than being rigidly determined by the content and style images}. This perspective necessitates a reevaluation of the role of training methodologies in style transfer processes, leading to the revelation that the generative model employed to describe the distribution of images has, in fact, already assimilated the nuanced art of transfer.

The emergence of diffusion models has catalyzed unprecedented interest in text-guided image manipulation and translation. These models demonstrate a remarkable capacity to generate artistic renditions when provided a style-related textual prompts. However, it is crucial to note that the efficacy of textual prompts is often limited by their inherent coarseness, which frequently proves inadequate in articulating the intricate details of desired stylistic elements.

Previous methodologies, such as InST\cite{zhang:2023:inversion} and VCT\cite{cheng:2023:ict}, have attempted to utilize image-controlled diffusion models for the purpose of image style transfer and translation. However, these approaches are constrained by the requirement to optimize a unique style embedding for each input style. This constraint presents significant challenges in the extraction and representation of precise stylistic features, often resulting in undesirable outcomes, produce results that deviate substantially from the intended input style while simultaneously failing to adequately preserve the original content (as illustrated in Fig.~\ref{fig:frame}(b)). 
Some researchers ~\cite{t2i-adapter,ipadapter} train a style-adapter network to extract style information of art painting and use the image-based style information as text prompt to ensure proper style pattern. However, it can not applied to specific content images. 
Based on IP-Adapter ~\cite{ipadapter}, InstantStyle-Plus ~\cite{instancestyle} introduces an image extractor for Content Preserving which can realize typical style transfer. However, it also faces the problem of content leakage and style deviation(as illustrated in Fig.~\ref{fig:frame}(c)).
In contrast to \cite{zhang:2023:inversion, cheng:2023:ict,instancestyle}, where control information is encoded as text embedding, we propose that the vanilla diffusion model is capable of extracting style information directly from the desired style image and fusing it into the content image \textit{without requiring re-training or tuning}.

In this paper, we leverage the style distribution from latent diffusion~\cite{rombach:2022:sd} and propose a \uline{z}ero-shot (\textit{i.e.,} training-free) \uline{s}tyle \uline{t}ransfer method via \uline{a}djuting style dist\uline{r}ibution, namely \textit{$ZSTAR^+$} to addresses the issues above. 
To obtain generative image priors, we employ dual diffusing paths to invert the style and content images. The features extracted from diffusion models inherently encapsulate both content and style information. Subsequently, we implement local style adjustment through an attention mechanism and global style adjustment via normalization operations.
For local style adjustment, we propose a Cross-attention Reweighting strategy that adeptly manipulates local content and style information from images and seamlessly fuses them within the diffusion latent space.
By leveraging a tailored attention mechanism, our diffusion model naturally addresses the constraints of content and style without necessitating additional supervision. We conduct an in-depth analysis of the style distribution differences between latent features of the style image and generated results during the denoising process. Our findings demonstrate that the proposed attention mechanism effectively aligns the style distribution, thus enhancing the fidelity of style transfer while maintaining content integrity.
However, even though the style distribution is closer, there is also a global style gap between the latent features of style and generated results, which will affect the color performance.
To address this issue, we propose a scaled adaptive instance normalization (SAIN) to adjust the global style distribution in the first denoising step while not affecting the content representation. 
Moreover, by considering the attention and energy loss between frame features, our method can be flexibly applied to video style transfer tasks.
Empirical evaluations of our proposed approach yield compelling outcomes, showcasing the method's capacity to produce high-quality results that effectively maintain the integrity of the original content while simultaneously incorporating vibrant stylistic elements that are harmoniously integrated with the underlying content structures.
In summary, our main contributions are as follows:
\begin{itemize}
\item  We propose a flexible image style transfer method by considering the local and global style distribution knowledge in the diffusion model to conduct image stylization without retraining/tuning.
\item  We propose a reweighted attention mechanism to perform local style rendering by disentangling and fusing content/style information in the diffusion latent space. 
\item  We propose the SAIN to adjust the global style distribution in the initial step, which is beneficial to better color alignment while not damaging the content information.
\item  We extend our method to video style transfer tasks. Various experiments demonstrate that our method can generate outstanding style transfer results, naturally fusing and balancing content and style from two input images. 
\end{itemize}

This work extends the conference paper “$\mathcal{Z}^*$: \uline{Z}ero-shot \uline{S}tyle \uline{T}ransfer via \uline{A}ttention \uline{R}eweighting” which
is published in CVPR2024~\cite{Deng_2024_CVPR}. 
Firstly, we extend the original method by considering the local and global style adjustment and reveal the style distribution in the latent space of diffusion model. 
Secondly, we propose a Scaled Adaptive Instance Normalization (SAIN) technique to better align the color distribution in stylized results with style images.
Furthermore, we demonstrate the applicability of our method to video-style transfer by leveraging inter-frame correlation.
We conduct extensive experiments and related analyses to elucidate the principles underlying the effectiveness of our method.

\section{Related Work}
\label{sec:related_work}

\subsection{Style transfer}
Style transfer aims to generate an image that retains the content layout from the input content while adopting a similar style to that of the input style.
With the development of machine learning, researchers design various neural networks to realize style transfer. 
The networks are trained by ensuring that the generated image and the content/style image exhibit content/style similarity~\cite{9994780, gatys:2016:image, Huang:2017:Arbitrary}.

Gatys \etal~\cite{gatys:2016:image} discover that Gram matrices of image features extracted from pre-trained VGG models could be used to represent style and propose an optimized-based method to generate stylized results by minimizing the Gram matrices differences. 
AdaIN ~\cite{Huang:2017:Arbitrary}  introduces an adaptive instance normalization technique to globally adjust the mean and variance of content images to align with style images.
Li et al.~\cite{li:2017:universal} employe a whitening and coloring transform (WCT) to align the covariance of style and content images, facilitating the transfer of multi-level style patterns and yielding enhanced stylized outcomes.
SANet~\cite{deng:2020:arbitrary} and MAST ~\cite{park:2019:arbitrary} leverage attention matrices between content and style features to imbue content with appropriate stylistic patterns.
AdaAttN~\cite{liu:2021:adaattn}  introduces an attentive normalization approach, applied on a per-point basis, to achieve feature distribution alignment.
InfoStyler~\cite{DBLP:journals/tcsv/LyuJPD24} focuses on disentangling the content and style features and introduces an information bottleneck to improve the feature representation ability for better stylization results.
ArtFlow~\cite{an:2021:artflow} proposes a flow-based network architecture designed to minimize image reconstruction errors and mitigate recovery bias.
Several recent studies~\cite{Deng:2022:CVPR, wu:2021:styleformer,wei:2022:comparative,Tang_2023_CVPR,wang2022fine,zhang2022s2wat} utilize the long-range feature represention ability of Transformer~\cite{xu:2022:transformers} to enhance stylization effects.
StyleFormer~\cite{wu:2021:styleformer}  incorporates the transformer mechanism into the traditional CNN-based encoder-decoder framework.
$StyTr^2$ addresses the biased representation issue inherent in CNN-based style transfer methodologies by employing a purely transformer-based architecture. However, the aforementioned approaches predominantly rely on second-order statistics of the entire image to quantify style information, which may prove insufficient for comprehensive style representation. Recent advancements~\cite{zhang:2022:domain,chen2021artistic,yang2023zero} have introduced contrastive loss as an alternative to the traditional Gram matrix-based style loss, demonstrating enhanced efficacy in processing fine-detail stylistic patterns.
CAST~\cite{zhang:2022:domain} innovatively considers the interrelationships between positive and negative examples, thereby addressing the limitation of existing style transfer models in fully utilizing extensive style information. Similarly,
IEST~\cite{chen:2021:iest} leverages the synergy of contrastive learning and external memory mechanisms to enhance visual quality.

Notwithstanding the continuous advancements in existing methodologies, the challenge of precise style representation persists. Inaccuracies in style expression can lead to suboptimal stylized outcomes. Acknowledging this ongoing challenge, our research aims to develop a zero-shot style transfer approach that circumvents the reliance on explicit style constraints, potentially offering a more robust and versatile solution to the style transfer task.

\subsection{Image generation based on diffusion model}
Diffusion models are popular in the generative research domain, which works by corrupting training data by continuously adding Gaussian noise and then learning to recover the data by reversing this noise process.
Diffusion models have demonstrated impressive results in text-to-image generation task~\cite{rombach:2022:sd,Ramesh:dalle2:2012,Saharia:imagen:2022,nichol2021glide}.
While certain methodologies, such as Imagic~\cite{Bahjat:imagic:2022} necessitate fine-tuning the entire diffusion model for each instruction, resulting in a process that is both time-consuming and computationally intensive. To mitigate these limitations, Prompt-to-prompt~\cite{Hertz:p2p:2023} replaces or reweights the attention map between text prompts and edited images, offering a more efficient alternative.
Additionally, NTI~\cite{Mokady:nulltext:2022} proposes null-text optimization based on Prompt-to-prompt, enabling real image editing capabilities. 
In order to reduce dependence on text prompts, StyleDiffuison~\cite{li2023stylediffusion} incorporates a mapping network to invert the input image into a context embedding, which is subsequently utilized as a key in the cross-attention layers.
Plug-and-Play~\cite{Tumanyan:plugplay:2022} and MasaCtrl~\cite{tecent:masactrl:2023}  shift focus to spatial features, leveraging self-attention mechanisms within the U-Net architecture of the latent diffusion model. 

While these methods demonstrate capability in text-guided style transfer through the use of prompts such as ``a pencil drawing'', simple textual descriptors may not suffice in capturing the intricacies of fine-detail style patterns.
Textual inversion~\cite{txtualinversion} can train a style embedding using a small group of style images.
T2I-adapter~\cite{t2i-adapter} and prompt-free diffusion~\cite{Prompt_Free} train a style extraction module to convert a style image to a context embedding to guide the generated result exactly.
The ControlNet team also proposed a “shuffle” model ~\cite{controlnet} to realize image-guided generation.
However, the content image can not be applied to control the semantics of the results.
To address this limitation, besides employing a style transfer/translation scheme that can train a style image into a style embedding to guide the generated results, InST~\cite{zhang:2023:inversion} and VCT~\cite{cheng:2023:ict} use the inversion process to maintain the input content semantics.

In this paper, we demonstrate that latent diffusion models can effectively perform image-guided style transfer using only style images, without the need for pseudo-text guidance or additional training. Our findings indicate that style images alone provide sufficient information for these models to achieve successful style transfer outcomes.

\section{Preliminary}
\label{sec:preliminary}
\subsection{Attention Mechanism}
The Attention Mechanism, initially introduced by Bahdanau et al.\cite{BahdanauCB14} as a powerful tool for information aggregation in neural network architectures, was subsequently adopted and refined by Vaswani et al.\cite{VaswaniSPUJGKP17} as a fundamental building block for machine translation. The core operation of the attention mechanism is defined as:

\begin{equation}
\text{Attention}(Q,K,V)=\text{Softmax}(\frac{QK^T}{\sqrt{d}})V.
\end{equation}

\subsection{Diffusion Model}
Diffusion Model is categorized as a generative model that leverages Gaussian noise for data sample generation. This model operates through a bi-directional process: a forward process and a reverse process. The reverse process, which is central to the model's generative capability, involves the gradual removal of this added noise. Conversely, the forward process systematically introduces noise to an initial data sample $x_0$, yielding a noisy sample $x_t$ at time-step $t$, in accordance with a predefined noise schedule $\alpha_t$:
\begin{equation}
    x_t=\sqrt{\alpha_t}\cdot x_0+\sqrt{1-\alpha_t}\cdot z,\text{ } z\sim\mathcal{N}(0,\mathbf{I}).
\label{forward_process}
\end{equation}
Additionally, a corresponding denoising diffusion 
model (DDIM) reverse process can also defined:

\begin{equation}
\begin{split}
    x_{t-1} &= \sqrt{\alpha_{t-1}} \cdot  \hat{x}_0^t \\
    &+  \sqrt{1-\alpha_{t-1} - \sigma_t^2 } \cdot \boldsymbol{\epsilon}_\theta(\mathbf{x}_t, t)    + \sigma_t z,  \\
     \hat{x}_0^t &= \frac{x_t-\sqrt{1-\alpha_t} \boldsymbol{\epsilon}^t_\theta(\mathbf{x}_t, t)}{\sqrt{\alpha_t}},
\end{split}
\label{ddim}
\end{equation}
where $\hat{x}_0^t$ is a prediction of the noise-free sample.
The backward process aims to gradually denoise $x_T\sim \mathcal{N}(0,\mathbf{I})$, where a cleaner image $x_{t-1}$ is obtained at each step. This is accomplished by a neural network $\boldsymbol{\epsilon}_\theta(x_t, t)$ that predicts the added noise $z$. 

A prevalent approach in contemporary research involves the utilization of a U-Net architecture integrated with an attention mechanism, serving as the noise prediction function $\boldsymbol{\epsilon}_\theta(x_t, t)$.
This configuration facilitates the implementation of self-attention to capture long-range dependencies among image features, while simultaneously employing cross-attention to incorporate guiding signals from the provided text prompt.
The attention mechanisms within this framework are formally expressed as:
\begin{equation}
    f_t^l=\text{Attention}(Q_t^l,K_t^l,V_t^l).
\label{attention_form}
\end{equation}
It is noteworthy that while the $K$ey and $V$alue at $l$-th layer may be derived from diverse sources, such as image spatial features or textual embeddings, they consistently adhere to the standardized attention formulation. This versatility allows for the seamless integration of multimodal information within the unified attention framework.
\section{Method}
\label{sec:Method}
\begin{figure*}
\setlength{\abovecaptionskip}{2mm}
\centering
\includegraphics[width=0.99\linewidth]{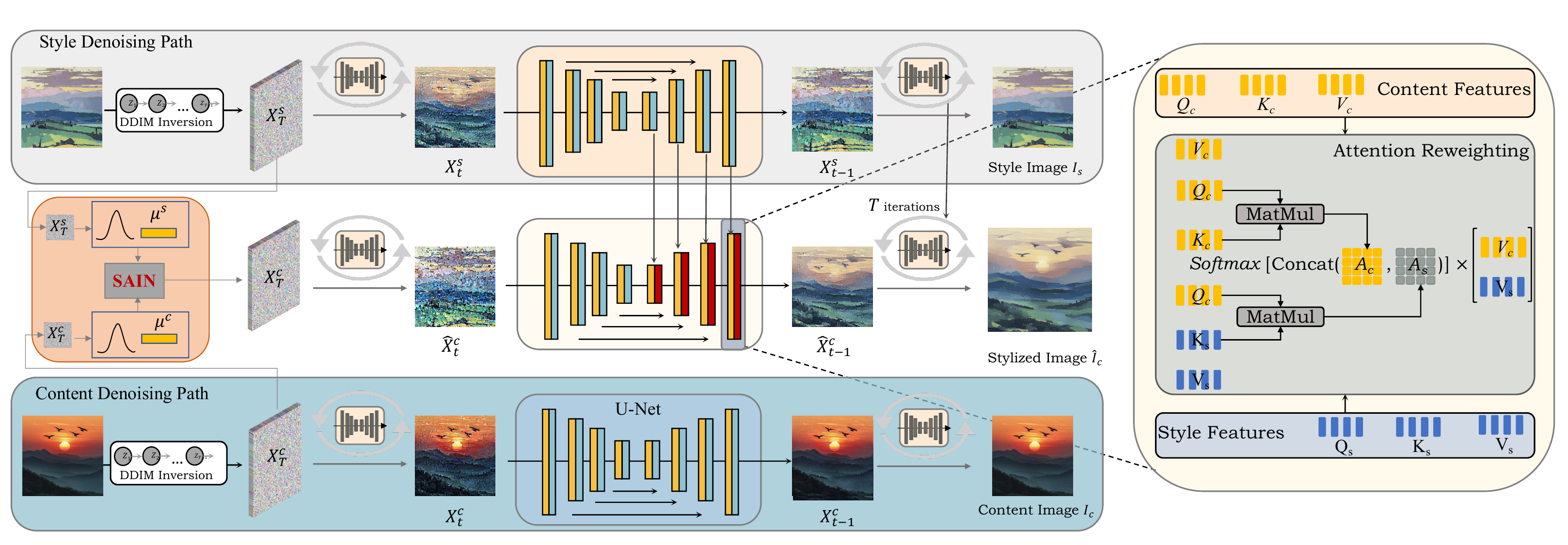}
\caption{Overall pipeline of our style transfer framework. The stylization process operates in the latent space. The process begins with separate DDIM inversions for both the content and style images. Firstly, we use the SAIN to adjust global style distribution in the initial step. During the subsequent denoising process, we employ our novel Cross-attention Reweighting technique to integrate local style patterns into the content structure. The stylization is achieved through an iterative process of 30 denoising steps, ultimately producing the final stylized output.
}
\label{fig:network}
\vspace{-10pt}
\end{figure*}

\subsection{Overview}
In this paper, we adopt local and global style distribution adjustment strategies in the pre-trained diffusion model.
The attention module within the stable diffusion can effectively align features $K$ and $V$ with the query $Q$. 
Previous studies, such as \cite{tecent:masactrl:2023, Tumanyan:plugplay:2022}, have leveraged the self-attention layer to extract information from key and value features, which represent local spatial attributes in the DDIM inversion process, for image editing applications.
Therefore, we design a dual-path network to generate $Q$, $K$ and $V$ features during the reverse process for content and style images respectively.
Then we propose a cross-attention reweighting technique to better align local content features with style features.
To close the style distribution between content and style feature globally, we analyze the style distribution trend in the latent space and propose a scale adaptive instance normalization to realize global feature adjustment.
At last, We also introduce video style transfer by considering the relationship between frame features.
\subsection{Dual-path Networks}
In our proposed methodology, we address a critical limitation inherent in conventional stable diffusion models. Specifically, we tackle the issue where text embeddings remain static throughout the reverse process, from timestamp $t\in[0,T]$, despite the inherent necessity for style features to adapt dynamically to the evolving denoised stylized image. This adaptation is paramount, given that the initial stages of denoising primarily involve reconstructing the image's fundamental shape and color, while the later stages focus on refining intricate details such as contours and brushstrokes.
To mitigate this limitation, we introduce an innovative dual-path framework that concurrently generates both the denoised style image and the stylized content image at an identical timestamp $T$. This is mathematically expressed as follows:
\begin{equation}
  I_s=\mathcal{G}_\theta(\epsilon_{I_s}, \{f_s\}, T),\quad I_c=\mathcal{G}_\theta(\epsilon_{I_c}, \{f_c\}, T).
\end{equation}
This approach ensures that the features in both networks are intrinsically aligned along the temporal dimension. Specifically, given a content image $I_c$ and a style image $I_s$, our objective is to derive a stylized output $\hat{I_c}$ that preserves the content of $I_c$ while incorporating the stylistic elements of $I_s$. This is achieved through the following formulation:
\begin{equation}
  \hat{I_c}=\mathcal{G}_\theta(\epsilon_{I_c}, \{f_s, f_c\}, T),
  \label{network_arch}
\end{equation}
where $\mathcal{G}_\theta(\cdot,\cdot,T)$ denotes the application of $T$ denoising steps in the diffusion model, utilizing fixed pre-trained weights $\theta$. The term $\epsilon{I_*}$ represents the noisy $x_T$ generated during the forward process by progressively introducing Gaussian noise to $I_c$ or $I_s$, as delineated in Eq.~(\ref{forward_process}). The notation $\{f_s, f_c\}$ refers to the spatial U-Net features derived from the style and content images.

As illustrated in Fig.~\ref{fig:network}, we employ DDIM inversion to invert both the style and content images, thereby obtaining the diffusion trajectories ${x}_{[0:T]}^c$ and ${x}_{[0:T]}^s$. Initially, we perform a global style distribution adjustment using our proposed SAIN module. Subsequently, we introduce a novel cross-attention arrangement to disentangle and fuse content and style information locally (i.e., $f_c$ and $f_s$, denoted as $Q$uery, $K$ey, and $V$alue) within the diffusion latent space using U-Net at timestamp $t$. Through $T$ denoising steps, we transform the stylized latent features $\hat{f}_{c}$, generated by the reweighted attention, into the final style transfer result $\hat{I}_{c}$.

\subsection{Local Style Adjustment}
\label{sec:method.attre}
\begin{figure}
\setlength{\abovecaptionskip}{2mm}
\centering
\includegraphics[width= \linewidth]{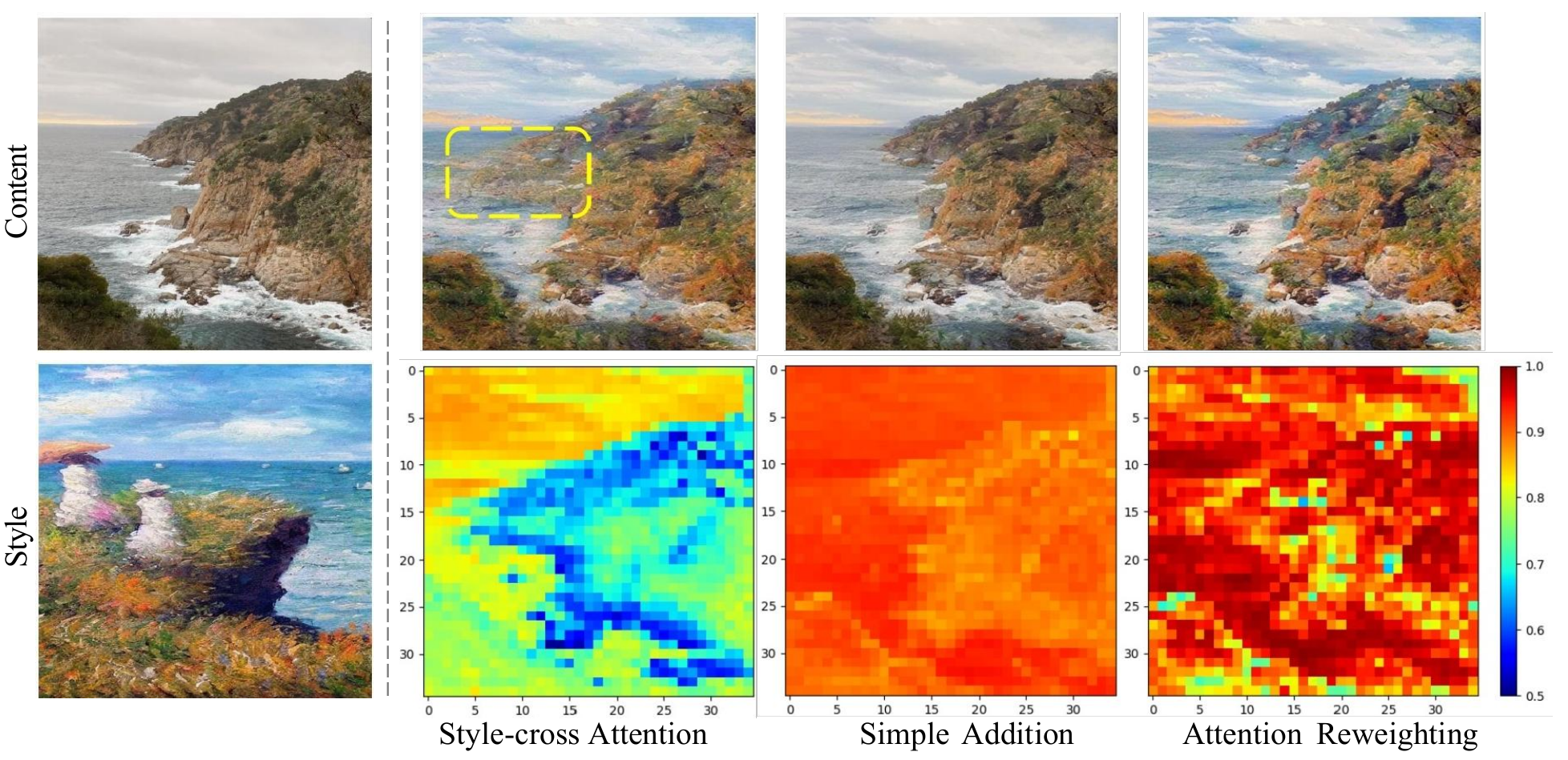}
\caption{The results obtained using style-cross attention demonstrate a tendency to overemphasize stylistic elements at the expense of preserving original content structures.Distortions are observed in areas where there is low correlation between self-attention and cross-attention outputs, indicating that $V_s$ inadequately reconstructs these target regions. ``Simple Addition'' retains an excessive amount of content features, whereas the Cross-attention Reweighting achieves a optimal trade-off.}
\label{fig:similarity}
\vspace{-10pt}
\end{figure}

To adjust the style distribution of the input content according to the target style in local fashion, we introduce the Cross-Attention Reweighting in this section.
Our attention mechanism incorporates two types of attention calculation between $f_s$ and $f_c$: style-cross attention for the integration of content and style features, and content self-attention for the preservation of structure. Since we use the standard self-attention, we primarily discuss the proposed style-cross attention in this section.
\paragraph{Style-cross Attention}
The style-cross attention mechanism utilizes content features to extract the most relevant information from style images for each input patch. It is natural to encode content information, such as the structure of an image, with the query ($Q$), while style information, encompassing attributes like color, texture, and object shape, is captured through the key ($K$) and value ($V$) features.  Formally, the inputs of style-cross attention are features from the content latent space $c$ and style latent space $s$ where 
\begin{equation}
    \hat{f}_c=\text{Attn}(Q_c,K_s,V_s)=\text{Softmax}(\frac{Q_cK_s^T}{\sqrt{d}})V_s.
\label{eq:naive_setting}
\end{equation}
Despite its apparent simplicity, empirical observations reveal that the naive fusion approach delineated in Eq.~(\ref{eq:naive_setting}) exhibits a propensity to overemphasize style patterns, often at the expense of preserving original content structures. Fig. \ref{fig:similarity}  illustrates this phenomenon through a heatmap depicting the cosine similarity between the outputs of cross-attention $\text{Atten}(Q_c,K_s,V_s)$ and self-attention $\text{Atten}(Q_c,K_c,V_c)$. 
The heatmap reveals a notable correlation between regions of low similarity scores and areas where content information has been significantly attenuated or lost. This observation suggests that the naive fusion approach may inadvertently prioritize style transfer over content preservation, particularly in regions where the style and content features exhibit substantial divergence.

\begin{figure}[t]
\begin{minipage}[t]{.15\textwidth}
  \centering
  \includegraphics[width=1.1\linewidth]{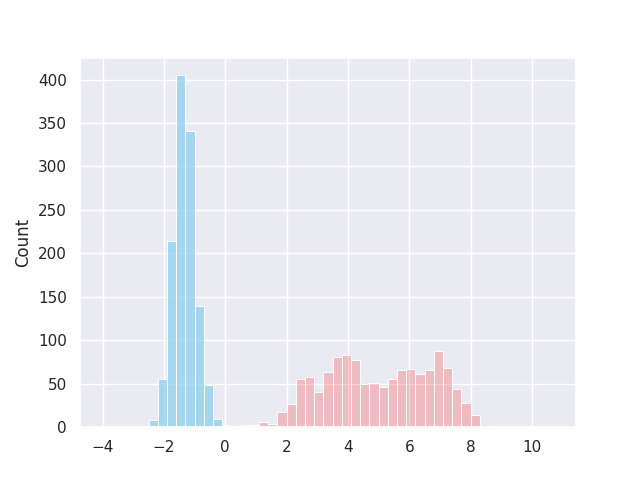}
  \caption*{(a) $\vec{q_c}K_s$}
\end{minipage}
\begin{minipage}[t]{.15\textwidth}
  \centering
  \includegraphics[width=1.1\linewidth]{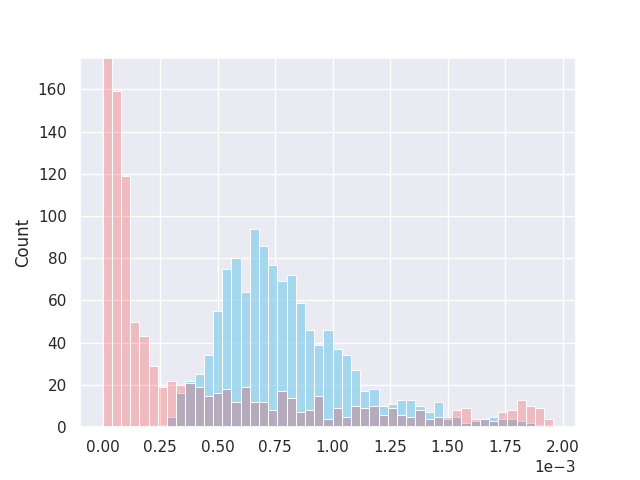}
  \caption*{(b) Softmax$(\vec{q_c}K_s)$}
\end{minipage}
\begin{minipage}[t]{.15\textwidth}
  \centering
  \includegraphics[width=1.1\linewidth]{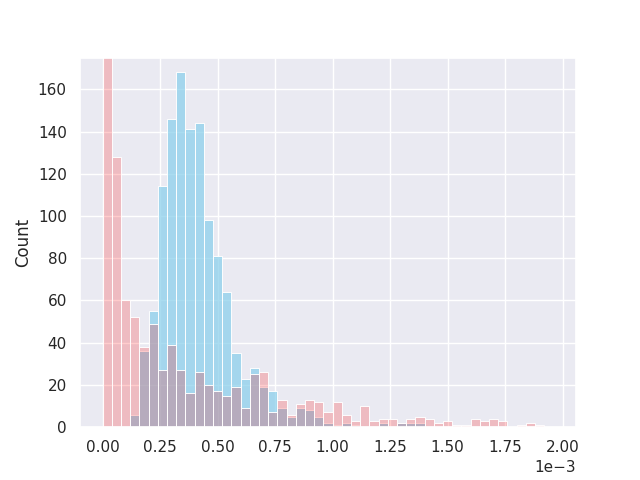}
  \caption*{(c) Ours}
\end{minipage}
\caption{Visualization of the distribution of $\vec{q_c}K_s$ for two content feature points (represented by blue and red bars) before and after the Softmax in (a) and (b). In (c), we represent the distribution of $\vec{q_c}K_s$ normalized by Eq.(\ref{ours_form}). A notable observation is that Softmax operation tends to overly amplify smaller values of $\vec{q_c}K_s$ (e.g., blue bars being shifted more significantly towards the right compared to the red bars in (b), despite the original $\vec{q_c}K_s$ values are mostly less than 0). In contrast, the negatively correlated values are limited to small magnitudes before and after normalization in (c).}
\vspace{-3mm}
\label{fig:histogram}
\end{figure}
\begin{figure}
\setlength{\abovecaptionskip}{2mm}
\centering
\includegraphics[width= \linewidth]{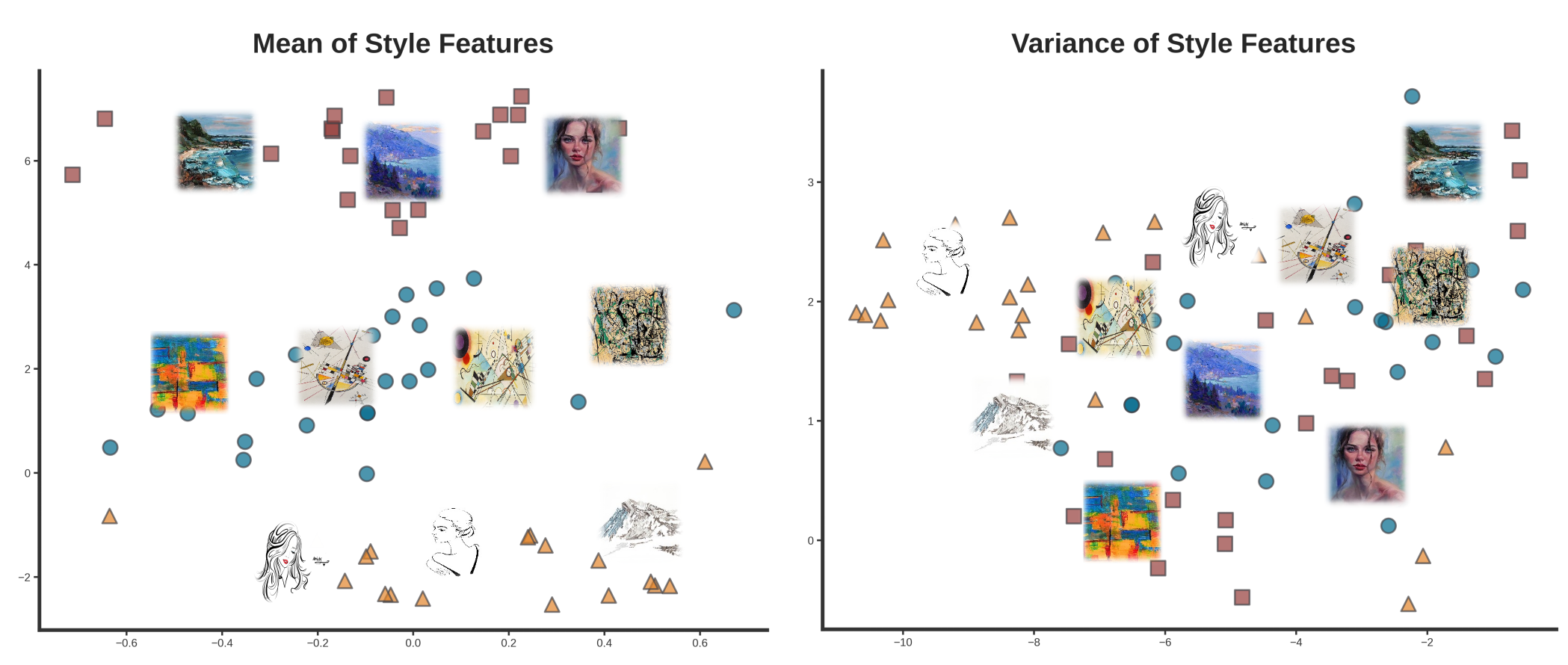}
\caption{The t-SNE visualization using mean and various latent features of a set of style images.
}
\label{fig:mean}
\end{figure}

\paragraph{Simple Addition}
To address the aforementioned challenge, we propose a straightforward solution that enhances the content information in $\hat{f}_c$ by reincorporating content self-attention. This approach is mathematically formulated as follows:
\begin{equation}
   \hat{f}_c=\lambda\cdot\text{Attn}(Q_c,K_s,V_s)+(1-\lambda)\cdot\text{Attn}(Q_c,K_c,V_c),
   \label{cross_attention_naive_add}
\end{equation}
where $\lambda\in[0,1]$.
We discovered that the selection of the $\lambda$ parameter requires careful consideration. This is particularly evident when examining the correlation between content pixels and style elements, as quantified by the values of $\vec{q}_cK_s^T$. Fig.~\ref{fig:histogram}(a) illustrates this relationship, where the blue bars represent instances of weak correlation (i.e., $\vec{q}_cK_s^T < 0$). To optimize the model's performance, it is advantageous to assign reduced attention weights to these pixels, thereby mitigating their potential negative influence.
However, the inherent properties of the Softmax function, which disregards absolute magnitudes and solely amplifies differences between $\vec{q}_cK_s^T$  values, lead to counter-intuitive outcomes.
As illustrated in Fig.~\ref{fig:histogram}(b), smaller $\vec{q}_cK_s^T$ values paradoxically result in larger attention weights after Softmax normalization. To mitigate this issue, we introduce an additional variable, $\lambda$, as a compensatory measure. However, it is crucial to note that a pre-defined $\lambda$ value cannot adequately address the diverse requirements of all content/style image pairs.
\paragraph{Cross-attention Reweighting}
While a handcrafted $\lambda$ external to $\text{Atten}(\cdot,\cdot,\cdot)$ can not adapt to input images, we discover that incorporating $\lambda$ within $\text{Softmax}(\cdot)$ achieves the desired adaptability. By reformulating Eq.~(\ref{cross_attention_naive_add}) in matrix form, we obtain:
\begin{equation}
\begin{split}
   \hat{f}_c&=\left[
   \begin{matrix}
       \lambda\cdot\sigma(\frac{Q_cK_s^T}{\sqrt{d}}), & (1-\lambda)\cdot\sigma(\frac{Q_cK_c^T}{\sqrt{d}})
   \end{matrix}\right] *\left[
   \begin{matrix}
       V_s\\
       V_c
       \end{matrix}\right]\label{former_atten_form}\\
   &=A*V'^T.
\end{split}
\end{equation}
Here, $\sigma(\cdot)$ denotes the Softmax function, and each row in $A\in\mathbb{R}^{N\times 2N}$, denoted as $\vec{a}\in\mathbb{R}^{2N}$, is normalized, \text{i.e.,} $\vec{a}\cdot \vec{\mathbf{1}}^T=1$. This normalization inspires us to re-construct the matrix $A$ in the form of applying Softmax to rows, i.e.,
\begin{align}
    A'&=\sigma(\left[
   \begin{matrix}
       \lambda\cdot\frac{Q_cK_s^T}{\sqrt{d}}, & \frac{Q_cK_c^T}{\sqrt{d}}
   \end{matrix}\right])\label{ours_form}\\
\hat{f}_c'&=A'*V'^T=\sigma(\left[
   \begin{matrix}
       \lambda\cdot\frac{Q_cK_s^T}{\sqrt{d}}, & \frac{Q_cK_c^T}{\sqrt{d}},
   \end{matrix}\right]) * \left[
   \begin{matrix}
       V_s\\
       V_c
\end{matrix}\right].
\label{lambda}
\end{align}
 Specifically, Eq.~(\ref{ours_form}) integrates $\lambda$ within Softmax, which inherently normalizes the output to the $[0,1]$ range, eliminating the requirement for $1-\lambda$.
In contrast to the previous attention formulation presented in Eq.~(\ref{former_atten_form}), the newly proposed reweighted attention matrix $A'\in\mathbb{R}^{N\times 2N}$ simultaneously considers both intra-content feature differences and inter-content-style feature differences during the application of the $\text{Softmax}(\cdot)$ function for output normalization. The reweighted attention matrix effectively enhances significant values of both $\vec{q}_cK_s^T$ and $\vec{q}_cK_s^T$ at each pixel, while automatically suppressing the small values of $\vec{q}_cK_s^T$ when the content pixel corresponding to $\vec{q}_c$ exhibits low relevance to all style pixels.

The Cross-attention Reweighting mechanism demonstrates enhanced flexibility and adaptability, characterized by several significant properties: (i) In scenarios where the correlation between the style and content images is weak, i.e., when each element $\vec{q}_c\vec{k_s}^T$ in $Q_cK_s^T$ approaches $-\infty$, the modified attention $\hat{f}_c'=A'*V'^T$ converges to the standard self-attention of content images, expressed as $\text{Attention}(Q_c,K_c,V_c)$. (ii) In instances of strong correlation between style and content images, where the maximum value of $\vec{q}_c\vec{k_s}^T$ approximates the maximum value of $\vec{q}_c\vec{k_c}^T$, and the Softmax operation generates an approximate one-hot probability distribution, then $\hat{f}_c'=A'*V'^T$ is equivalent to Eq.~(\ref{cross_attention_naive_add}). (iii) Notably, Eq.~(\ref{cross_attention_naive_add}) can be reformulated  using $A'$ as follows:
\begin{equation}
\begin{split}
    \hat{f}_c&=\frac{1}{2}\cdot\text{Attn}(Q_c,K_s,V_s)+\frac{1}{2}\cdot\text{Attn}(Q_c,K_c,V_c),\\
    &=\sigma(\left[
   \begin{matrix}
       \frac{Q_cK_s^T}{\sqrt{d}}+C, & \frac{Q_cK_c^T}{\sqrt{d}}
   \end{matrix}\right])*\left[
   \begin{matrix}
       V_s\\
       V_c
   \end{matrix}\right],
   \label{naive_add_new_form}   
\end{split}
\end{equation}
where
\begin{equation}
    C=\ln{\frac{\sum_j\exp{\big([Q_cK_c^T]_{\cdot,j}\big)}}{\sum_j\exp{\big([Q_cK_s^T]_{\cdot,j}\big)}}}.
\end{equation}
In Eq.~(\ref{naive_add_new_form}), the simple addition of self-attention to cross-attention introduces an additional term $C$. This variable serves to magnify all elements within $Q_cK_s^T$, including small values that represent weak correlations between style and content features, which are typically considered inconsequential and should ideally be disregarded. Consequently, the incorporation of $C$ may introduce an increased level of noise into the Softmax function.
This noise introduction can potentially lead to sub-optimal outcomes in the stylization process, highlighting a limitation of the simple addition approach in effectively balancing self-attention and cross-attention for style transfer tasks.

\begin{figure}
\setlength{\abovecaptionskip}{2mm}
\centering
\includegraphics[width= \linewidth]{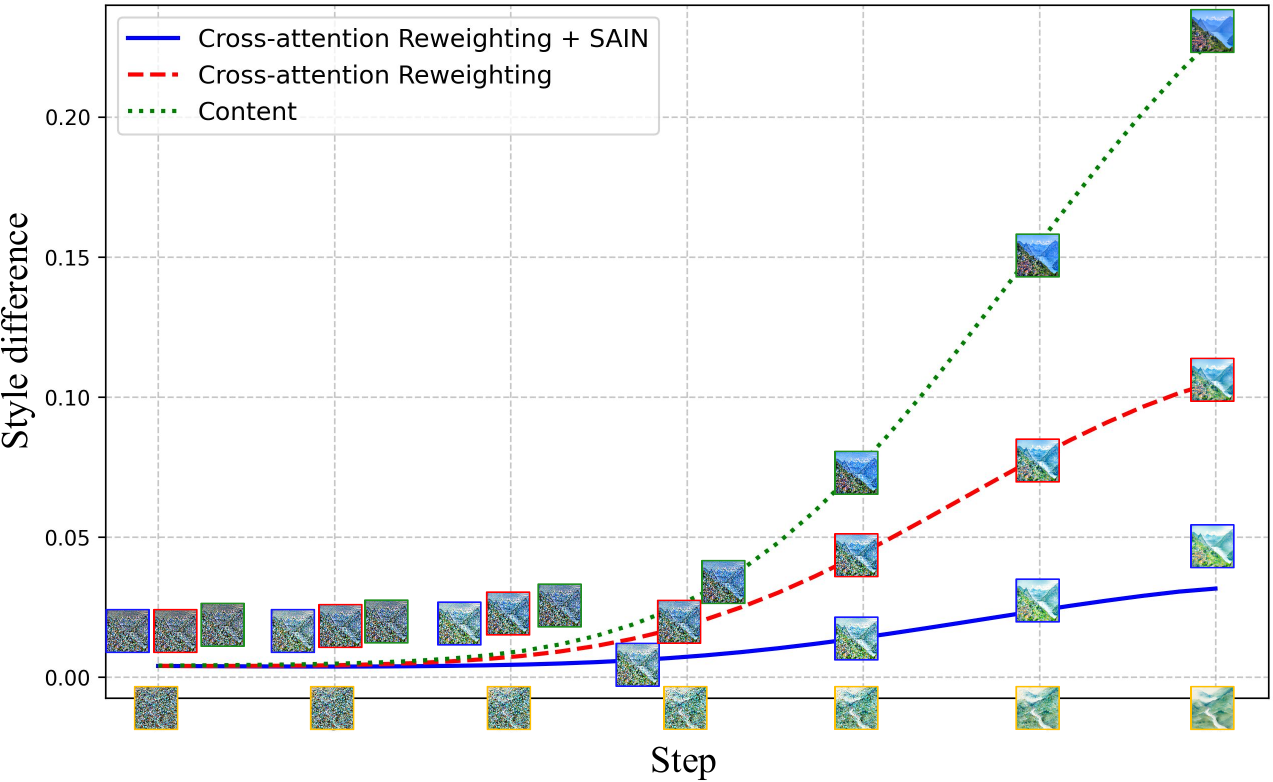}
\caption{Style difference between stylized/content features and style features in the denoising process. The images enclosed in yellow boxes represent visualized style features, while those in green boxes depict visualized content features. Images within red boxes showcase stylized features obtained through local style adjustment. The images contained in blue boxes demonstrate stylized features resulting from both local and global adjustments..
}
\label{fig:style_diff}
\end{figure}

\subsection{Global Style Adjustment}
\label{sec:style_distribution}
The Cross-attention Reweighting module enables effective disentanglement and fusion of content and style information. This module utilizes local content features to query the most suitable information from style images for each input patch by computing attention between content and style features. This operation significantly narrows the gap between the style distribution of generated results and that of style images.
We calculate the style differences between style features and content/stylized features in different steps in diffusion models using Eq.~(\ref{style_loss}).
\begin{equation}
\begin{split}
\mathcal{L}^t_{style} &= \lVert G_t(\hat{I}_{c}) - G_t(I_{s})  \lVert_2 ,\\
G_t(x) &= \frac{1}{C_t H_t W_t} \sum_{h=1}^{H_t}\sum_{w=1}^{W_t} \phi_t (x)_{h,w,c}\phi_t (x)_{h,w,c'}
\end{split}
\label{style_loss}
\end{equation}
where $\phi_t(\cdot)$ denotes the features extracted from the $t$-th step in the diffusion model, $G_t(\cdot)$ is proportional to the uncentered covariance of the extracted features.

We visualize the feature in each step during the generation process and trends in style differences in  Fig.~\ref{fig:style_diff}.
As the number of denoising steps increases, the style difference also increases due to the emergence of more visual information.
Our Cross-attention Reweighting module effectively reduces the style difference between style features and stylized features compared to the difference between style features and content features. This observation demonstrates the module's capability to transfer style patterns from the style image to the content image.
Although the style distribution is narrowed by the cross-attention reweighting, we neglect that the global content and style features have deviations in style distribution which will affect the attention between them.
This deviation causes the content image's color to influence the stylization result, potentially weakening the stylization effect.
This observation raises a crucial question: how can we reduce global differences and further eliminate inconsistencies in color distribution?

See Fig.~\ref{fig:style_diff}, in the early step, the features are almost pure noise. 
The style loss is small, but there is still a slight difference.
We assume that the implicit style is determined by the noise feature obtained by DDIM inversion process. Adjusting the style distribution in the initial can gradually eliminate inconsistencies in color distribution between style and stylized image.

To test this hypothesis, we select a group of style images across 3 styles and use the t-SNE to visualize the variance and mean of initial noise features.
As shown in Fig.~\ref{fig:mean}, the mean of initial noise features is gathered according to different styles, while the variances show less style-specific agglomeration.
This confirms our hypothesis: implicit style is determined by the mean of noise feature obtained by DDIM inversion process.
Therefore, we can adjust the style distribution of stylized features in the initial step to align the results and style image better.

\paragraph{Scaled adaptive instance normalization}


Directly, we apply Eq.~(\ref{mean}) in the initial step to replace the mean of the content features with the mean of the style features.

\begin{equation}
\begin{split}
\hat{f}_{c} &= f_c - \mu_c + \mu_s
\end{split}
\label{mean}
\end{equation}
However, given that initial noise is highly correlated with the final content results, this mean adjustment operation may risk disrupting the input content. To address this challenge, we propose Scaled Adaptive Instance Normalization (SAIN), which dynamically determines the degree of manipulation by measuring the distributional differences between the style and content images.
\begin{equation}
\begin{split}
\hat{f}_{c} &= f_c - \mu_c * w  + \mu_s * w
\end{split}
\label{adain}
\end{equation}
where $w$ are determined by the K-L divergence between initial content feature and style feature:
\begin{equation}
\begin{split}
w = e^{ KL(f_s||f_c)}  \\
KL(f_s||f_c) = \sum  p(f_s) \log \frac{p(f_s)}{q(f_c)}
\end{split}
\label{kl}
\end{equation}
The notation $p(f_s)$ denotes the probability density function of $f_s$, estimated via histogram-based methods, and analogously, $p(f_c)$ represents the probability density function of $f_c$ derived through the same approach.
When the probability distribution difference between initial content and style is lager, the scale wight are smaller.
Ensure that content is not destroyed by limiting the extent of distribution shift.
As shown in Fig.~\ref{fig:style_diff}, with SAIN, the style difference further decreases.
The color distribution of stylized results are better aligned with style images.

\subsection{Video style transfer}
\begin{figure}
\setlength{\abovecaptionskip}{2mm}
\centering
\includegraphics[width= \linewidth]{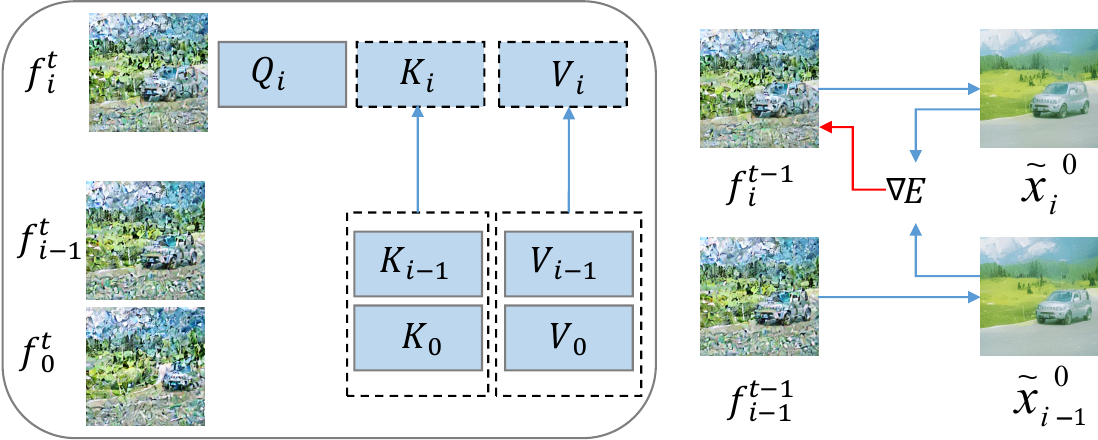}
\caption{Injecting attention module in different denoising steps and visualization for grad backward in the noise-free domain.
}
\label{fig:video_fream}
\end{figure}

Firstly, we extend our method to video style transfer by incorporating inter-frame attention.
At each diffusion step, we obtain the latent features of the current frame $f_i$ guided by the latent of the reference frame. We choose the first frame and the $i-1$ frame $f_{i-1}$ to increase the consistence of the stylized video.
As illustrated in Fig.~\ref{fig:video_fream}, we modify the attention mechanism for frame $f_i$ with by replacing its Key and Value with the concatenation of the corresponding Key and Value from frames $f_{i-1}$ and $f_0$. This approach enables the current frame to leverage information from both its immediate predecessor and the initial frame, promoting stylistic coherence throughout the video sequence.
\begin{equation}
\begin{split}
 K_i = \mathrm{Concat}(K_0, K_{i-1}),  V_i = \mathrm{Concat}(V_0, V_{i-1}) \\  
\end{split}
\end{equation}

While the inter-frame attention mechanism promotes video consistency, its effect is relatively subtle when based on an text-to-image diffusion model.
Inspired by ~\cite{Pix2Video},  we conduct consistency constrain in the noise-free domain to eliminate the flikering in the generated video farther.
We define a emergy loss defined as $E_{t-1} = \lVert \hat{x}^0_{i,t-1}  -  \hat{x}^0_{i-1,t-1}\lVert_2$ to minimize the  difference between consecutive frames $i$ and $i-1$ at each denosing step in the noise-free space, where $\hat{x}^0_{i,t-1}$ is predicted at each step using Eq.~(\ref{ddim})
Subsequently, we update predicted sample $f^i_{t-1}$ by minimizes $E$:
\begin{equation}
f^i_{t-1} = f^i_{t-1} - w \cdot \nabla_{f_t^i} \cdot E_{t-1}
\end{equation}

\begin{table*}
\small
\centering
\caption{User study results. Each number represents the percentage of votes that the corresponding method is preferred to ours, using  the criteria of overall quality, preservation of content and style, respectively.}
\resizebox{0.99\linewidth}{!}{
\begin{tabular}{c|cccccccc}
\toprule
  &  InstantStyle-Plus~\cite{instancestyle}& VCT~\cite{cheng:2023:ict}&InST~\cite{zhang:2023:inversion}&QuanArt~\cite{huang2023quantart}&CAST~\cite{zhang:2022:domain}&StyTr$^2$~\cite{Deng:2022:CVPR} & IEST~\cite{chen:2021:iest} &AdaAttN~\cite{liu:2021:adaattn}  \\
\midrule
content & 14.5&31.5\%& 29.5\%&45.5\%&35.5\%&45.0\% & 46.5\% & 32.0\%\\
\midrule
style  & 35.3& 33.0\% &  25.0\%&21.5\%&49.0\%&47.5\% & 43.5\% & 40.5\% \\
\midrule
overall &  30.0&27.5\% &  19.0\%&30.5\%&31.5\%&36.5\% & 35.5\% & 34.0\%\\
\bottomrule
\end{tabular}}
\vspace{-10pt}
\label{tab:quancomp}
\end{table*}

\begin{table*}
\small
\centering
\caption{Quantitative comparisons. To evaluate the preservation of input content and style, we calculate the average values of content and style loss for the results obtained through various methods. The most favorable outcomes are highlighted in \textbf{bold}.}
\resizebox{\linewidth}{!}{
\begin{tabular}{c|ccccccccc}
\toprule
  &ours&  InstantStyle-Plus~\cite{instancestyle}& VCT~\cite{cheng:2023:ict}&InST~\cite{zhang:2023:inversion}&QuanArt~\cite{huang2023quantart}&CAST~\cite{zhang:2022:domain}&StyTr$^2$~\cite{Deng:2022:CVPR} & IEST~\cite{chen:2021:iest} &AdaAttN~\cite{liu:2021:adaattn}  \\
\midrule
$\contentloss$ &  1.65&2.11&1.73 &  2.85 & 1.71 & 1.66 &  \textbf{1.59} &  1.82&1.89 \\
\midrule
$\styleloss$  & \textbf{3.63} &4.18&3.98&4.70 & 5.50& 4.18&3.79   & 3.86&3.78\\
\bottomrule
\end{tabular}
}
\label{tab:quancomps}
\end{table*}

\section{Experiments}
\label{sec:Experiments}

\subsection{Implementation Details}
Our study builds upon the concept of Stable Diffusion~\cite{RombachBLE:sd:2O22} and employs the v1.5 checkpoint of the model.
In our experimental protocol, we configure the text prompts as null character strings to isolate the visual influence to style transfer process. The denoising procedure is structured to encompass a total of $30$ steps.
Notably, we implement our novel cross-attention reweighting module between layers layers $10$-$15$ during the $5^{th}$-$30^{th}$ denoising steps. 
\subsection{Evaluation}
In this section, we compare our methods with  state-of-the-art style transfer approaches AdaAttN~\cite{liu:2021:adaattn}, IEST~\cite{chen:2021:iest}, StyTr$^2$~\cite{Deng:2022:CVPR}, CAST~\cite{zhang:2022:domain}, QuanArt~\cite{huang2023quantart}, InST~\cite{zhang:2023:inversion}, VCT~\cite{cheng:2023:ict} and InstantStyle-Plus~\cite{instancestyle}.

\begin{figure*}
\setlength{\abovecaptionskip}{2mm}
\centering
\includegraphics[width= 0.90\linewidth]{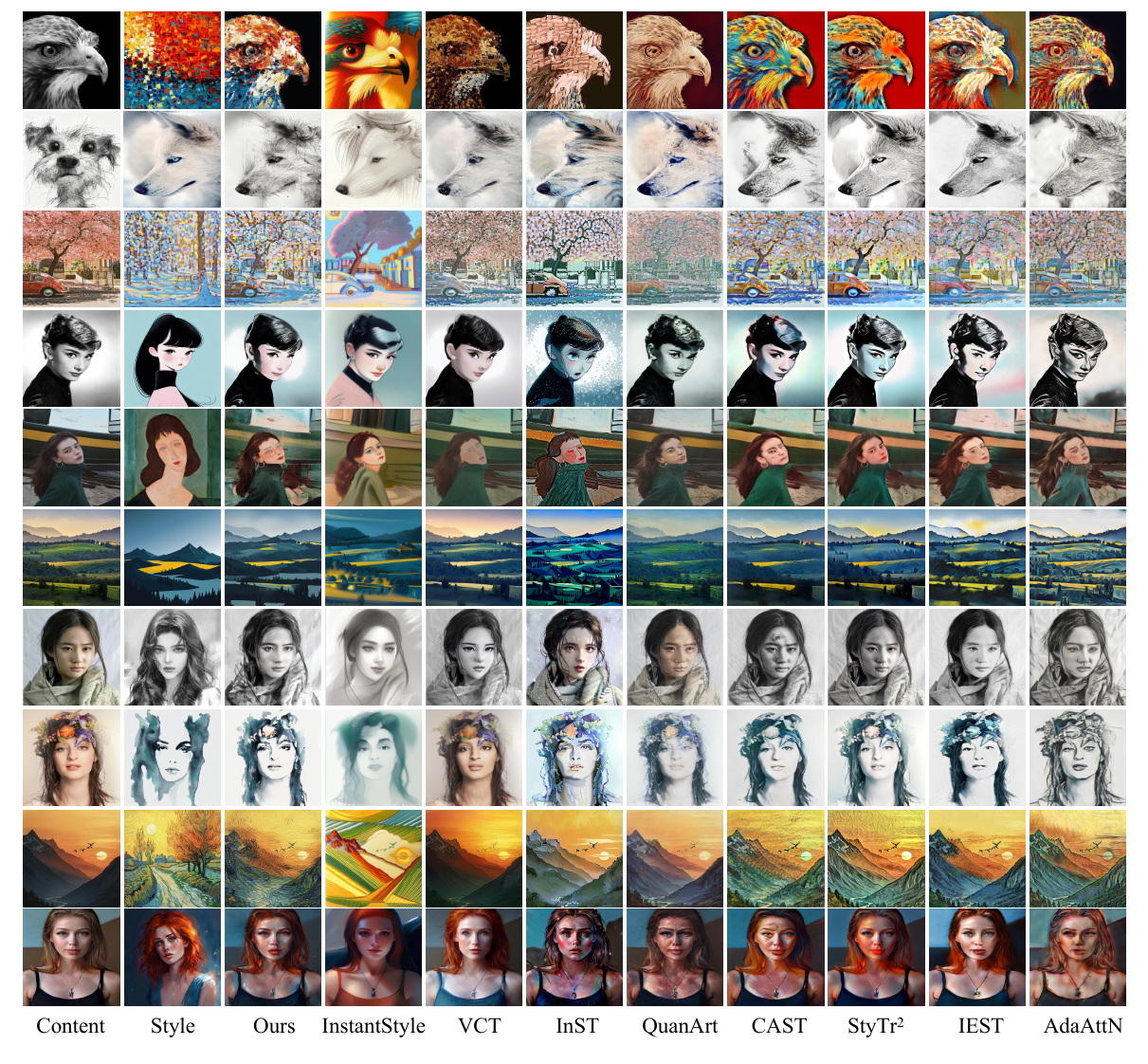}
\caption{Compared with other style transfer methods. The content images are presented in first column, the style images are presented in second column, and the stylized results generated by different methods are presented in the rest. 
}
\vspace{-2mm}
\label{fig:compare}
\end{figure*}

\paragraph{Qualitative evaluation}
The visual outcomes of various style transfer methods are qualitatively analyzed in Fig.~\ref{fig:compare}.
AdaAttN~\cite{liu:2021:adaattn} struggles to capture the intended style patterns particularly evident in the \engordnumber{4} ,\engordnumber{6} and \engordnumber{8} rows.
IEST~\cite{chen:2021:iest} demonstrates improved style rendering, it still exhibits discrepancies between the generated output and the input style reference (the \engordnumber{1} and \engordnumber{4} rows).
CAST~\cite{zhang:2022:domain} and StyTr$^2$~\cite{Deng:2022:CVPR} fall short in accurately reproducing artistic qualities, appearing ``artificial in comparison to genuine artworks'' due to inadequate style loss constraints (the \engordnumber{1}, \engordnumber{2} and \engordnumber{4} rows). 
The results of QuanArt~\cite{huang2023quantart} results display a diminished style performance, attributed to an overemphasis on content preservation.
InST ~\cite{zhang:2023:inversion} attempts to optimize style embedding but fails to generate outputs that faithfully represent both the input style and content (the \engordnumber{7} and \engordnumber{8} rows).
VCT~\cite{cheng:2023:ict} has enhanced content performance but continues to struggle with style bias issues (the \engordnumber{1} and \engordnumber{3} rows).
InstantStyle-Plus~\cite{instancestyle} try to learn the intrinsic content and style presentation, but introduce the content bias and style deviation (the \engordnumber{9} row and alteration of the facial identity in \engordnumber{7}).

In comparison, our proposed method achieves visually appealing stylized results by effectively transferring style elements, such as brush strokes and linear patterns, onto the input content images. These style elements are adeptly adapted to the semantic content, as exemplified in the \engordnumber{1}, \engordnumber{2}, and \engordnumber{7} rows of Fig.~\ref{fig:compare}.

\paragraph{User study}
To objectively assess the effectiveness of various stylization methods, we conducted a comprehensive user study to gauge public opinion. 
The study involved a diverse group of $50$ randomly selected participants.
For the evaluation, we utilized $10$ content images and $10$ style images, generating a total of $100$ stylized outputs using our proposed method and several comparison methods.
Each participant was presented with $32$ sets of questions. In each set, they were shown a randomly selected content/style image pair along with its corresponding stylized results from both our method and a randomly chosen comparison method.
Participants were asked to evaluate the stylized outputs based on three criteria: 
1) which stylization result better preserves the content, 
2) which stylization result exhibits better style patterns and 
3) which stylization result has a better overall effect.
The study yielded a total of $4,800$ individual votes.
The results of this voting process are summarized in Table~\ref{tab:quancomp}. 
Analysis of the data reveals that our method consistently outperforms all comparison methods in terms of content preservation and style representation.
Notably, our method demonstrates superior performance in overall effect, surpassing all comparative methods. This indicates that our method successfully achieves an optimal balance between style transfer and content preservation, resulting in visually pleasing outcomes.

\paragraph{Quantitative comparisons}
To conduct a quantitative assessment, our evaluation framework incorporates two key metrics: content loss and style loss. These measures serve to quantify the effectiveness of content preservation and style rendering, respectively, allowing for a comprehensive comparison of performance across different methodologies.
Content loss and style loss are commonly adopted in style transfer tasks, defined as:
\begin{equation}
\begin{split}
\contentloss = \frac{1}{\layernumber}\sum_{ i=0 }^{\layernumber} \lVert \phi_i(\hat{I}_{c}) - \phi_i(\inputcontent)  \lVert_2,
\end{split}
\label{fun:contentloss}
\end{equation}
where $\phi_i(\cdot)$ denotes features extracted from the $i$-th layer in a pre-trained VGG19 and $\layernumber$ is the number of layers. The style perceptual loss $\styleloss$ is defined as
\begin{equation}
\begin{split}
\styleloss = \frac{1}{\layernumber}\sum_{ i=0 }^{\layernumber}& \lVert \mu( \phi_i(\hat{I}_{c})) - \mu (\phi_i(I_{s})) \lVert_2 \\
&+ \lVert \sigma( \phi_i(\hat{I}_{c})) - \sigma (\phi_i(I_{s})) \lVert_2,
\end{split}
\label{fun:styleloss}
\end{equation}
where $\mu(\cdot)$ and $\sigma (\cdot)$ denote the mean and variance of extracted features, respectively. The content loss is computed using the features extracted from the $5$-th layer, while the style loss is calculated using the features from the $1$-st to $5$-th layers.

Table~\ref{tab:quancomps} presents the results of our comparative analysis. Our proposed approach exhibits the lowest style loss, indicating its exceptional capability in style rendering. Conversely, StyTr$^2$~\cite{Deng:2022:CVPR} achieves the lowest content loss, highlighting its effectiveness in preserving content. A notable aspect of our method is its ability to achieve comparatively low values for both content and style loss metrics, suggesting a favorable balance between content preservation and style rendering.

\begin{figure*}
\setlength{\abovecaptionskip}{2mm}
\centering
\includegraphics[width= 0.9\linewidth]{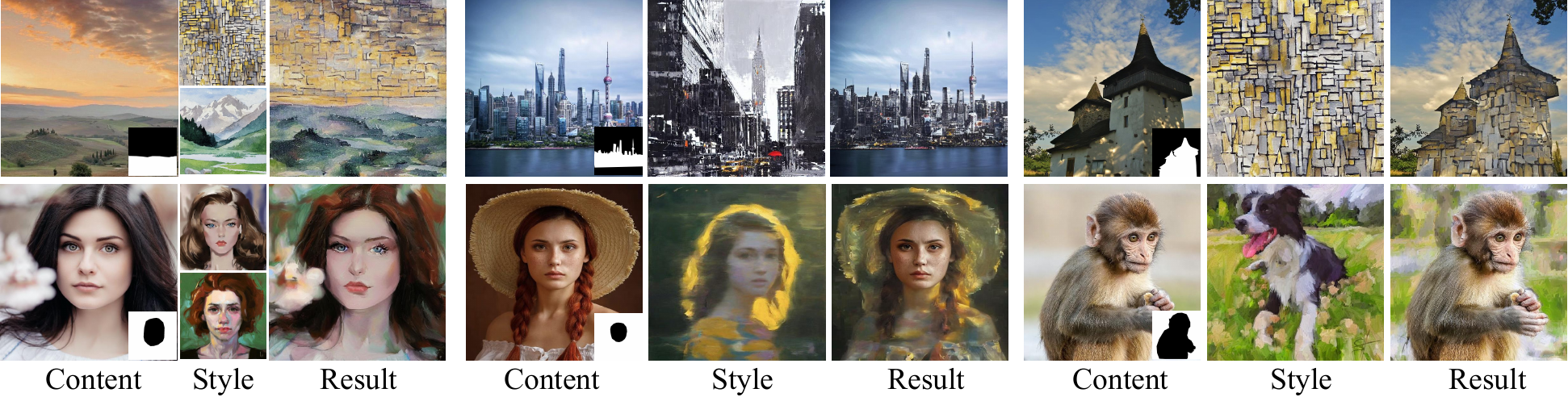}
\caption{The illustrations of regional control. The control map is represented by the binary mask located at the bottom right of the content reference. 
The first column represents multi-style regional control results, black pixels of the control map mark the areas designated for style transfer using the first style image, while white pixels of the control map mark the areas designated for style transfer using the second style image.
The rest columns represent single-style regional control results. White pixels mark the areas designated for style transfer.}
\label{fig:style_grad}
\vspace{-10pt}
\end{figure*}
\subsection{Conditional Control}
Our method is potential for extension to more sophisticated downstream applications, notably including conditional control utilizing binary masks. To facilitate this expansion, we introduce an auxiliary mapping function $\phi(\cdot)$ on $\frac{Q_cK_c^T}{\sqrt{d}}$, which enables more precise control over a designated region $\Omega$ during the image style transfer process. The augmented formulation can be expressed as:
\begin{equation}
A'=\sigma(\left[\begin{matrix}
       \phi(\frac{Q_cK_s^T}{\sqrt{d}}), & \frac{Q_cK_c^T}{\sqrt{d}}
   \end{matrix}\right]).
\end{equation}
Here, $\phi(x_{i,j})$ is defined as:
\begin{equation}
    \phi(x_{i,j})=
\left\{
\begin{array}{ll}
   -\infty & \{i,j\}\notin\Omega \\
   x_{i,j} & \text{otherwise}
\end{array}.
\right.
\end{equation}
It is crucial to acknowledge that the direct assignment of $-\infty$ values may result in abrupt style transitions, potentially creating undesirable artificial boundaries. To mitigate this issue and achieve a more naturalistic gradient effect, we implement a linear gradient function for $\phi(x_{i,j})$, which smoothly interpolates from $-\infty$ to $x_{i,j}$. 

Moreover, the proposed framework demonstrates considerable flexibility, allowing for a straightforward extension from the conventional one-to-one content-style image pair to a one-to-many configuration. This extension is readily achievable through a modification of Eq.~(\ref{ours_form}). Specifically, when aiming to transfer the styles of $N$ style images to a single content image, the equation is modified as follows:
\begin{equation}
    A'=\sigma(\left[
   \begin{matrix}
       \frac{Q_cK_{s_1}^T}{\sqrt{d}}, & ..., & \frac{Q_cK_{s_N}^T}{\sqrt{d}}, & \frac{Q_cK_c^T}{\sqrt{d}},
   \end{matrix}\right]).
\end{equation}
The effectiveness of our conditional control approach is visually demonstrated in Fig.~\ref{fig:style_grad}. The leftmost column presents results obtained using the one-to-two content-style image configuration, showcasing the method's capability to blend multiple styles. The subsequent columns display outcomes from the conventional one-to-one content-style image setting, providing a basis for comparison. Notably, our conditional control strategy enables the generation of stylized results featuring smooth transitions between distinct regions. This gradual blending of styles across different areas of the image underscores the sophisticated nature of our approach, allowing for nuanced and aesthetically pleasing style transfers that avoid abrupt or artificial-looking boundaries.

\begin{figure*}
\setlength{\abovecaptionskip}{2mm}
\centering
\includegraphics[width= 0.9\linewidth]{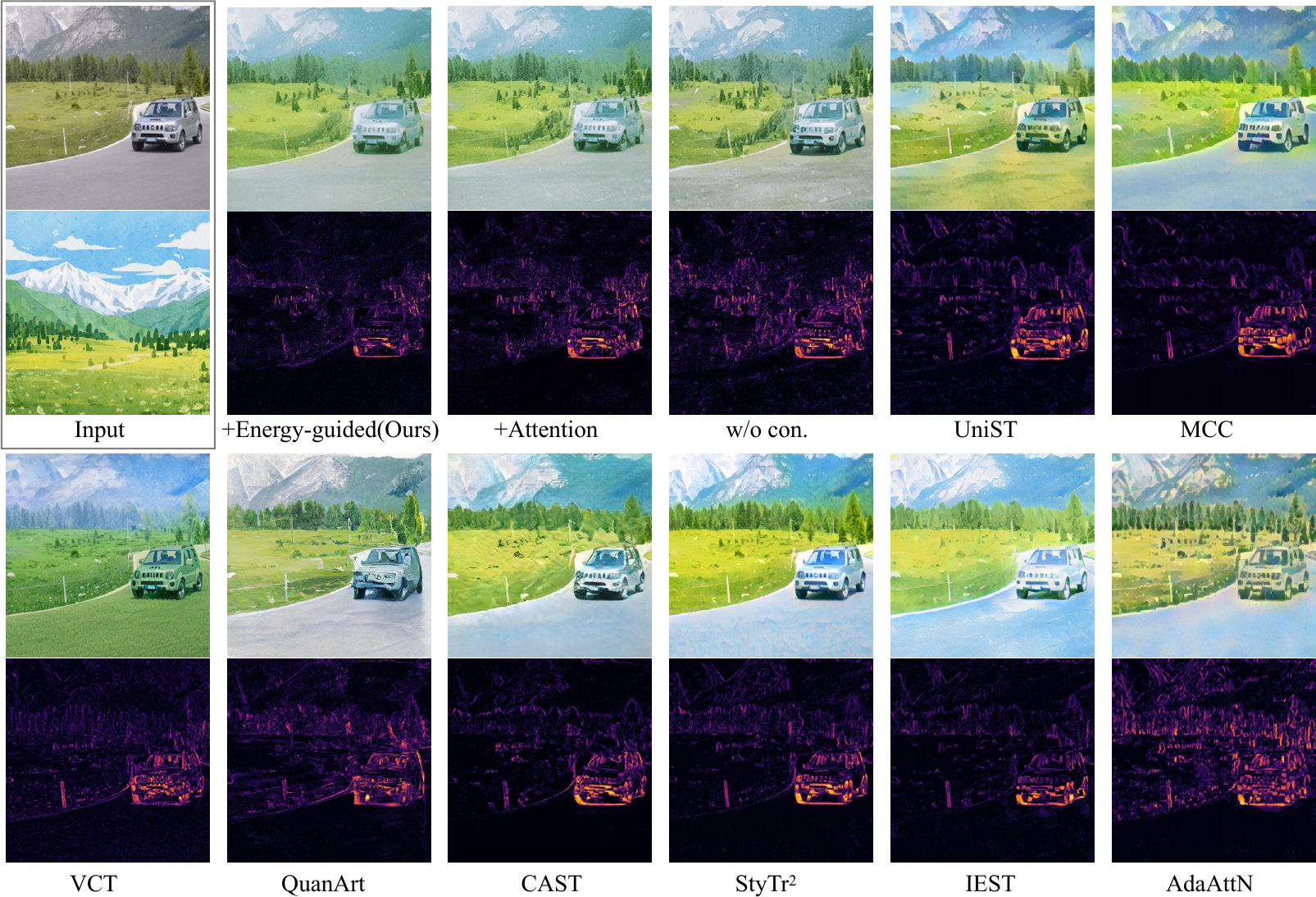}
\caption{Comparison of video style transfer results.
}
\label{fig:video_compare}
\end{figure*}

\subsection{Video Study Transfer}
We conducted a comprehensive comparative analysis of our proposed method against eight baseline approaches, including state-of-the-art video style transfer techniques such as UniST~\cite{Unist} and MCC~\cite{mcc2}.
We show the visual style transfer results in Fig.~\ref{fig:video_compare} and the difference between two adjacent stylized frames.
Without continuity constraints(w/o con. ), our initial image style transfer method faces the problem of video flikering when compared with contrast methods.
By considering the the attention of $i$-th frame with $i-1$-th frame and first frames, the consistency of stylized results are promoted that slightly exceeded the baseline method, such as IniST~\cite{Unist}, CAST~\cite{zhang:2022:domain} and IEST~\cite{chen:2021:iest}.
Further refinement was achieved through the minimization of energy loss, which substantially reduced inter-frame differences. This approach yielded a marked improvement in temporal stability.

We also introduce the $Diff=\lVert F_i - F_{i-1} \lVert $ and calculate the $mean(mean_{Diff})$ and $variance(var_{Diff})$ to Quantitatively evaluate the consistency performance.
Table~\ref{tab:video} presents the quantitative results of our consistency evaluation. The data demonstrates that our method consistently produces superior video results with high temporal coherence, outperforming the baseline approaches across the defined metrics.
\begin{table*}
\small
\centering
\caption{User study results. Each number represents the percentage of votes that the corresponding method is preferred to ours, using  the criteria of overall quality, preservation of content and style, respectively.}
\resizebox{\linewidth}{!}{
\begin{tabular}{cccccccccccc}
\toprule
 & Ours&+Attention&w/o con.& UniST~\cite{Unist} & MCC~\cite{mcc2}&VCT~\cite{cheng:2023:ict}& QuanArt~\cite{huang2023quantart}&CAST~\cite{zhang:2022:domain}&StyTr$^2$~\cite{Deng:2022:CVPR} & IEST~\cite{chen:2021:iest} &AdaAttN~\cite{liu:2021:adaattn}\\
\midrule
Mean & \textbf{0.0362}&0.0507&0.0723& 0.0536& 0.0491&0.0561&0.0718&0.0552 & 0.0605& 0.0519&0.0711\\
\midrule
Var.  & \textbf{0.0041} & 0.0051&0.0086& 0.0072&  0.0075&0.0040&0.0054&0.0068 &0.0055 & 0.0048&0.0066 \\
\bottomrule
\end{tabular}}
\vspace{-10pt}
\label{tab:video}
\end{table*}

\begin{figure}
\setlength{\abovecaptionskip}{2mm}
\centering
\includegraphics[width= 0.9 \linewidth]{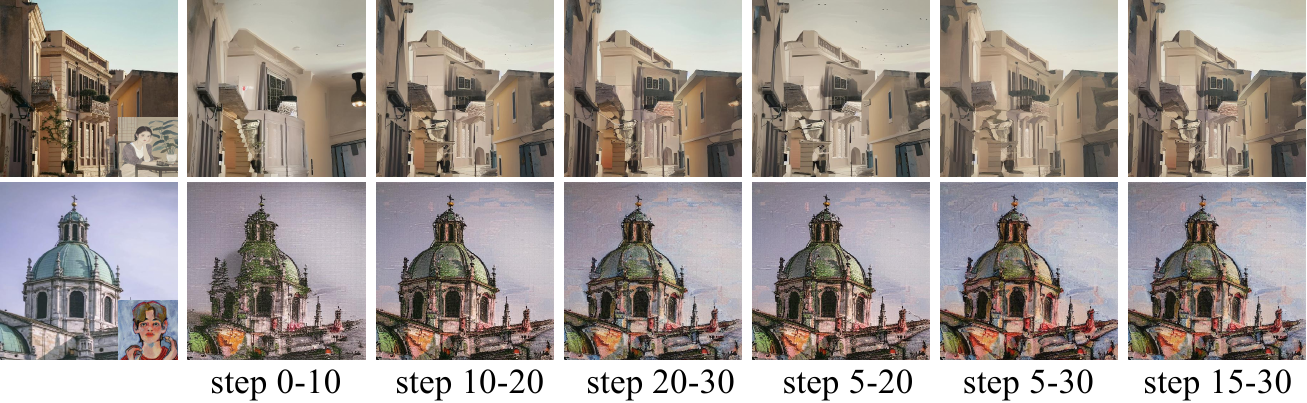}
\caption{Injecting attention module in different denoising steps.
}
\label{fig:a1}
\end{figure}

\begin{figure}
\setlength{\abovecaptionskip}{2mm}
\centering
\includegraphics[width= 0.85\linewidth]{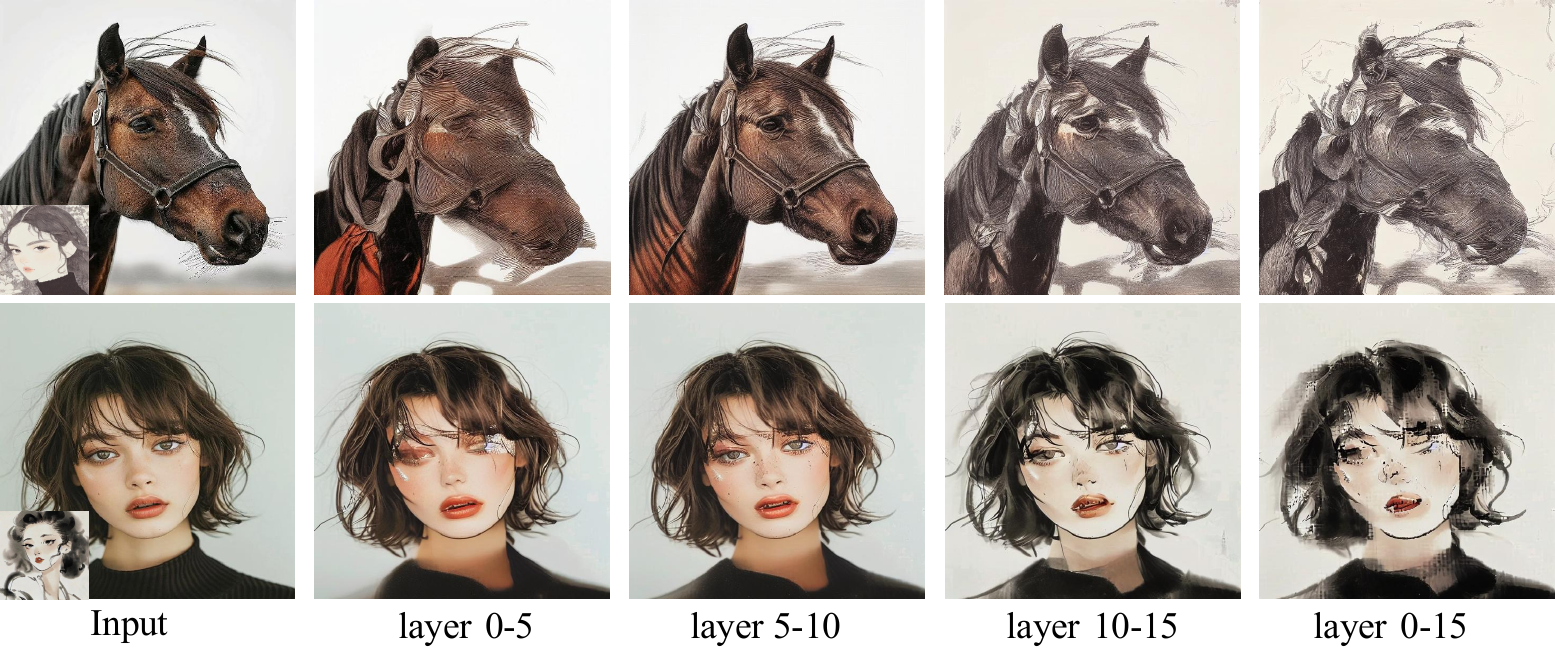}
\caption{Injecting attention module in different U-Net layers.
}
\label{fig:a2}
\end{figure}

\subsection{Ablation Study}
We conducted a series of comprehensive ablation studies to elucidate the individual contributions of each component to the overall efficacy of our style transfer methodology including the attention injection step/layer, cross-attention reweighting module, SAIN operation and the style scaling factor. Table~\ref{tab:ablation} presents a comprehensive quantitative comparison, offering insights into the relative impact of each component on the style transfer quality.
The following sections provide a detailed qualitative analysis of each element, complementing the quantitative results and offering a more nuanced understanding of their respective roles in the style transfer process.
\paragraph{Influence of the attention injection step}
The analysis focuses on the impact of the attention injection step on stylization outcomes.
As illustrated in Fig.~\ref{fig:a1}, the denoising step initiation significantly influences the stylization process. Commencing the denoising step prematurely results in a loss of content structure information, while delaying the start enhances content preservation (see the first three columns in Fig.~\ref{fig:a1}). Conversely, extending the overall number of denoising steps accentuates style patterns without compromising content structure, as evidenced by the comparison between the fourth and fifth result columns.
Furthermore, as discussed in Sec.~\ref{sec:style_distribution}, an increase in the number of denoising steps correlates with the emergence of more style visuals, contributing to more effective style rendering.
\begin{figure}
\setlength{\abovecaptionskip}{2mm}
\centering
\includegraphics[width= 0.9\linewidth]{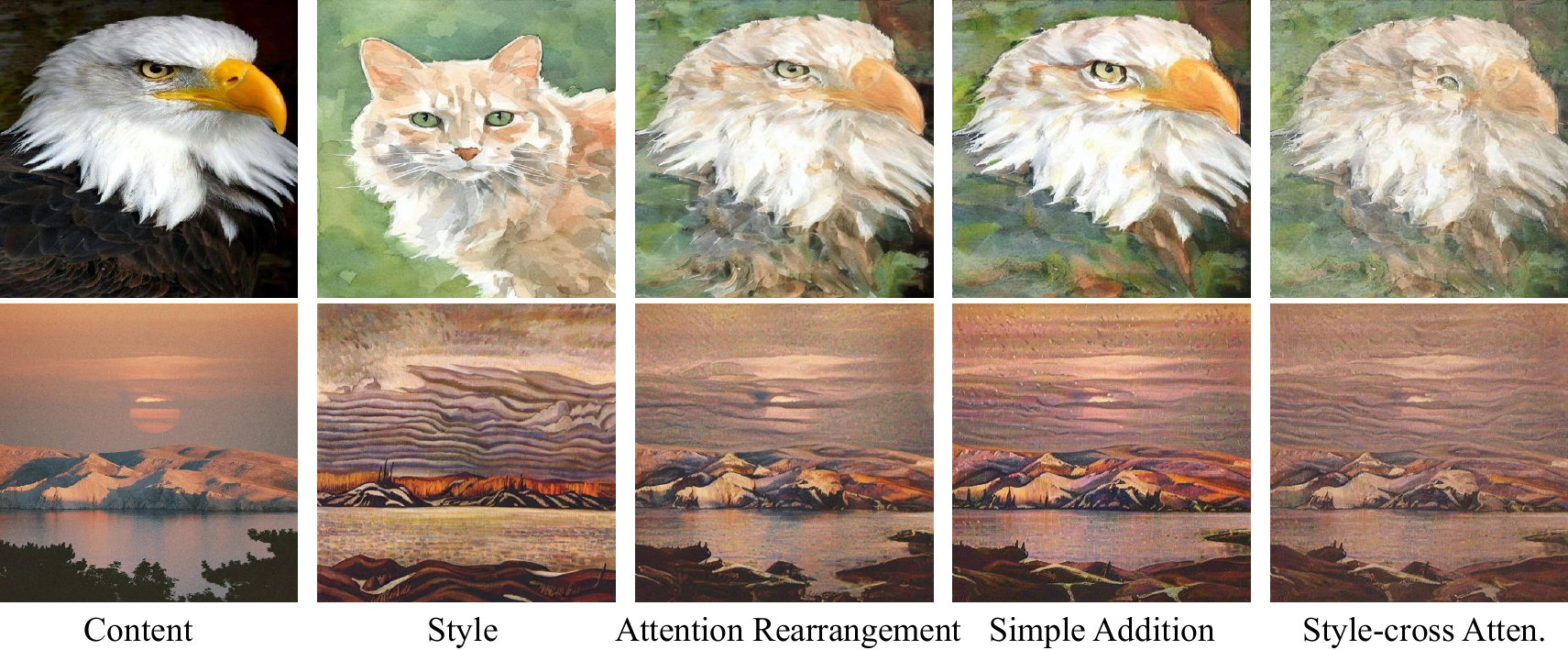}
\caption{Results of using different attention arrangement.
}
\label{fig:attention}
\vspace{-10pt}
\end{figure}
\begin{figure}
\setlength{\abovecaptionskip}{2mm}
\centering
\includegraphics[width= 0.9\linewidth]{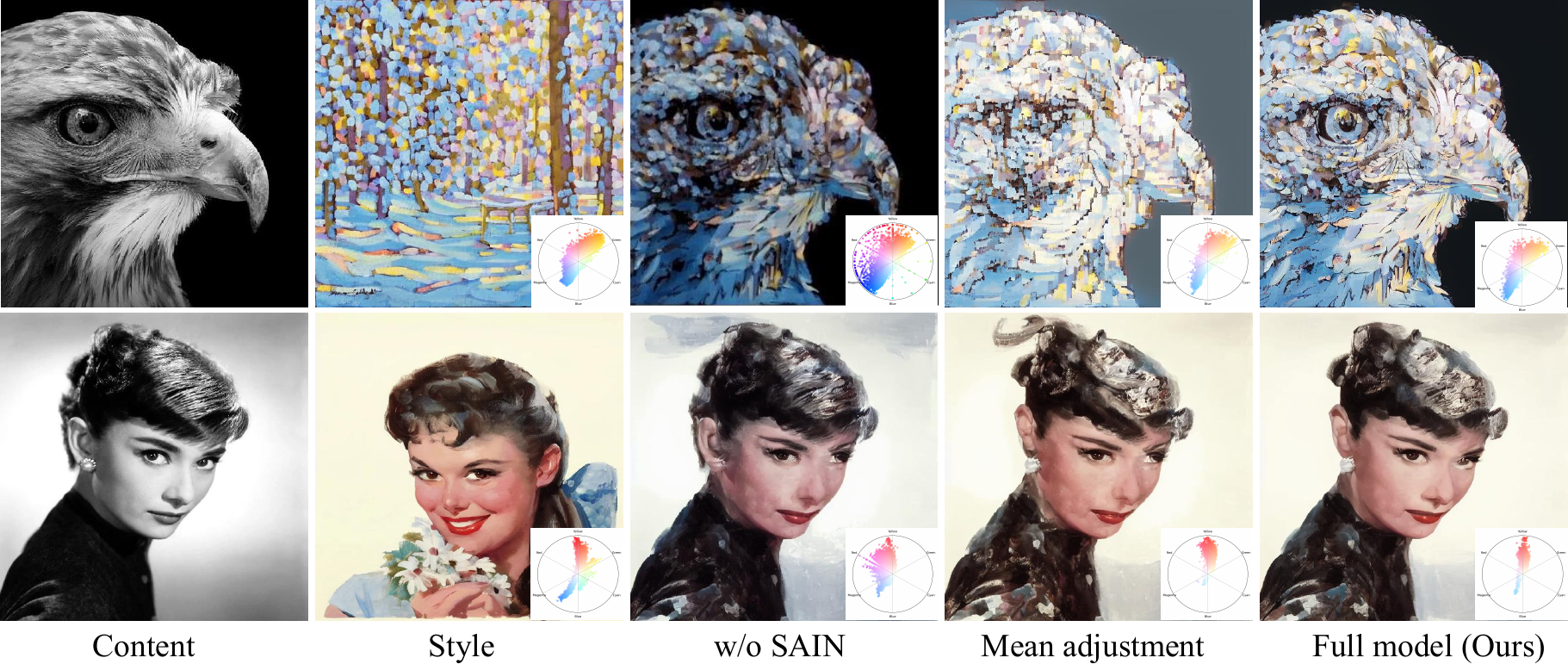}
\caption{Influence of the SAIN.
}
\label{fig:sain}
\end{figure}

Based on these observations, the optimal stylization results are achieved by initiating the denoising process at the \nth{5} step and concluding it at the \nth{30} step. This configuration strikes an ideal balance between preserving content structure and effectively incorporating style elements.
\paragraph{Influence of the attention injection layers}
\Cref{fig:a2} illustrates the stylized outputs resulting from the integration of the attention module at various layers of the U-Net architecture. This experimental analysis reveals the differential impacts of attention mechanisms across the network's hierarchical structure.
The indiscriminate application of attention mechanisms across all U-Net layers (layers 0-15) adversely affects the structural integrity of the content in the resultant images. A more nuanced examination reveals that the low-resolution layers (layers 5-10) of the U-Net architecture effectively preserve content structure, albeit at the cost of limited style pattern transfer. In contrast, the high-resolution layers (layers 0-5 and 10-15) exhibit a superior capacity for style feature extraction. However, it is noteworthy that the high-resolution layers within the encoder section (layers 0-5) tend to compromise content structure and weaken style patterns.

Based on these observations, we conclude that the optimal strategy involves the selective injection of the attention module into the high-resolution layers of the decoder. This approach facilitates effective style transfer while maintaining the structural integrity of the content.

\paragraph{Influence of the cross-attention reweighting}
To evaluate the efficacy of the cross-attention reweighting, we provide a visual representation of the stylized outcome in Fig.~\ref{fig:attention}. 
We conduct a comprehensive ablation study to evaluate the efficacy of our proposed approach. Two key scenarios are examined: 1) the complete removal of content, as described in Eq.~(\ref{eq:naive_setting}), and 2) the utilization of only the summation operation, denoted by Eq.~(\ref{cross_attention_naive_add}). 
For the latter, we assign a value of 0.5 to the coefficient $\lambda$.

As evidenced in Fig.~\ref{fig:attention}, the elimination of the content component results in a significant degradation of content structural preservation in the stylized outputs. Similarly, the sole incorporation of the Query feature derived from the content image leads to a loss of content fidelity and inadequate style pattern transfer. In contrast, our proposed cross-attention reweighting mechanism demonstrates superior performance in achieving an optimal balance between content preservation and style transfer in the output images.
\paragraph{Influence of the SAIN}
The SAIN is proposed to align global style distributions and mitigate color inconsistencies. To assess its effectiveness, we conduct experiments where the SAIN is either removed or replaced with a simple mean adjustment technique.
\begin{figure}
\setlength{\abovecaptionskip}{2mm}
\centering
\includegraphics[width= \linewidth]{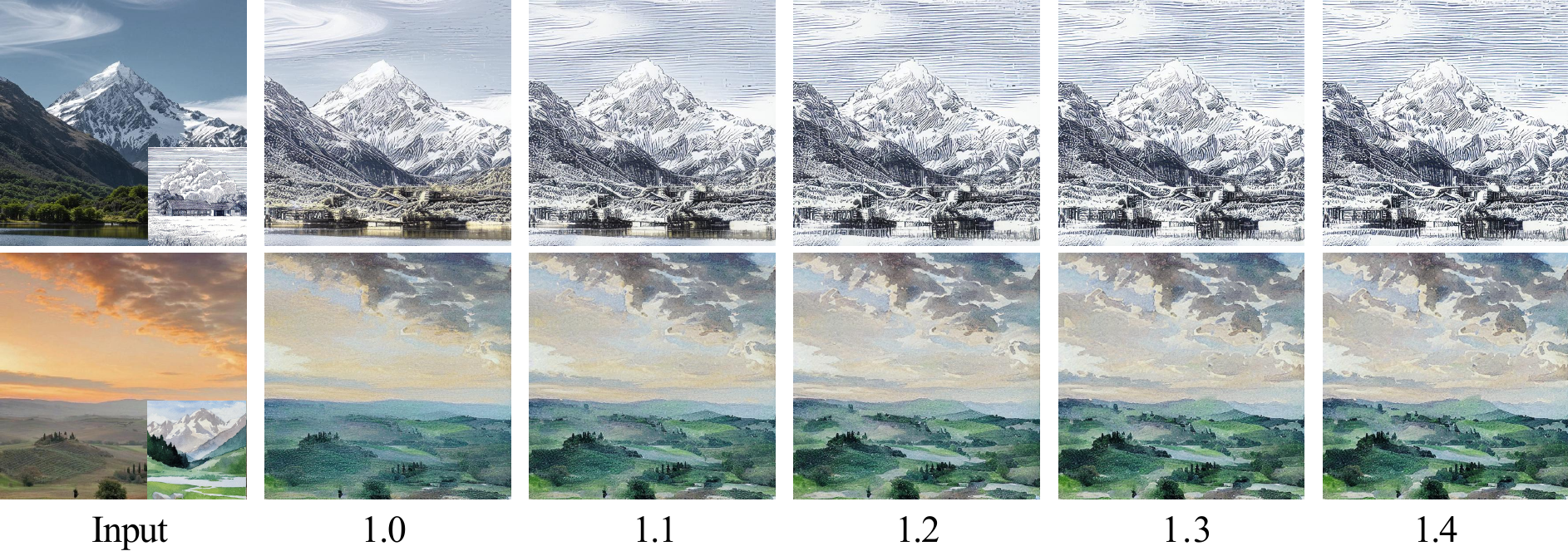}
\caption{The effect of $\lambda$ in Eq.~(\ref{lambda}) .
}
\label{fig:lambda}
\vspace{-10pt}
\end{figure}
\begin{table}
\tiny
\fontsize{3pt}{4pt}
\tabcolsep=0.5cm
\centering
\caption{Quantitative results for ablation study.}
\resizebox{0.9\linewidth}{!}{
\begin{tabular}{ccc}
\toprule
 &$\contentloss$ &$\styleloss$ \\
 \midrule
    ours & 1.65  & 3.63     \\
\midrule
    step 0-10     &  2.63&    4.98   \\
    step 10-20     &  1.19     &  4.48 \\
    step 20-30     &  1.13    & 4.96  \\
    step 5-20     &   1.42  &  4.01 \\
  step 15-30     &  1.34  &  4.47 \\
\midrule
      layer 0-5     & 2.06 &   5.07\\
    layer 5-10     &   1.11 &  6.05 \\  
layer 0-15     &    2.57 &  3.76
 \\
\midrule
    Simple Add &  1.78 &   4.11       \\
    style-across   & 1.93 &      3.55     \\
\midrule
    w/o SAIN &  1.61 &   3.70       \\
    Mean adjustment   & 1.88 &      3.59     \\
\midrule
    $\lambda$ = 1.0    &    1.13 &  4.61 \\
    $\lambda$ = 1.1     &    1.47 &  4.13\\
    $\lambda$ = 1.3     &  2.23  & 3.49\\
    $\lambda$ = 1.4     &    2.29 &  3.45 \\
    \bottomrule

\end{tabular}
}
\label{tab:ablation}
\end{table}

We show the stylized results in Fig.~\ref{fig:sain} and place the color wheel in the lower right corner of the corresponding image to show the color distribution. As illustrated in Fig.~\ref{fig:sain}, the integration of SAIN results in stylized outputs with coherent color distributions that closely align with the target style. Conversely, the absence of SAIN leads to noticeable deviations from the desired style in terms of color fidelity. Substituting SAIN with a simple mean adjustment technique adversely affects content structure, underscoring the efficacy of our adaptive scaling approach using Kullback-Leibler divergence in preserving content details while achieving style transfer.
\paragraph{Influence of the style scaling factor}
We conduct a comprehensive visual analysis to evaluate the influence of varying the parameter $\lambda$ in Equation (\ref{lambda}). The results of this analysis are presented in Fig. \ref{fig:lambda}, which illustrates the impact of different $\lambda$ values on the stylization outcomes.
We discover the robustness of the proposed method across a spectrum of $\lambda$ values. Notably, the approach exhibits consistent and satisfactory performance for $\lambda$ values greater than or equal to 1.2. 
Based on these findings, we adopt a fixed value of $\lambda = 1.2$ for all ensuing experiments in this study.

\section{Conclusion}
This study introduces a novel zero-shot style transfer approach that adjusts the local and global style distribution in the latent space of a pre-trained diffusion model. We propose a Cross-attention Reweighting module as a replacement for the conventional self-attention mechanism in diffusion models for local style adjustment. This innovative module enables the utilization of local Query features to extract style-related information from the Key and Value features, thereby enhancing the style transfer process. Furthermore, we introduce a scaled adaptive instance normalization technique to modulate the initial global style distribution while mitigating content leakage issues. Our method extends to video style transfer by considering correction between frame features, demonstrating compatibility with both image and video diffusion models. Comprehensive experimental evaluations substantiate the superiority of our proposed approach in terms of stylization quality, surpassing current state-of-the-art methods in the field.

{
\small
\bibliographystyle{IEEEtran}
\bibliography{ZST}
}

\end{document}